%% file: main_arxiv.tex
\documentclass[acmtog, nonacm]{acmart}

\setcopyright{none}
\renewcommand\footnotetextcopyrightpermission[1]{}
\input{paper_info}
\input{macros}

\usepackage{booktabs} 
\usepackage{subcaption}

\usepackage[ruled]{algorithm2e} 

\SetAlFnt{\small}
\SetAlCapFnt{\small}
\SetAlCapNameFnt{\small}
\SetAlCapHSkip{0pt}

\usepackage{duckuments}  

\begin{document}
\title{\ourPaperTitle}

\input{Main/Metadata/authors}
\input{Main/Sections/abstract}

\input{Main/Metadata/ccs}

%
%
\input{Main/Partials/figure_teaser}
\maketitle
\input{Main/Sections/intro}
\input{Main/Sections/related_work}
\input{Main/Sections/method}
\input{Main/Sections/experiment}
\input{Main/Sections/conclusion}
\clearpage

\bibliographystyle{ACM-Reference-Format}
\bibliography{main}

\clearpage
\input{Main/Sections/figure_only}
\clearpage
\appendix
\input{Supple/supple}

\end{document}

%% file: paper_info.tex
\def\ourPaperTitle {ATSplat: Compact Feed-forward 3D Gaussian Splatting \\with Adaptive Token Expansion}

%% file: macros.tex
\newcommand{\tref}[1]{Tab.~\ref{#1}}

\newcommand{\fref}[1]{Fig.~\ref{#1}}

\newcommand{\sref}[1]{Sec.~\ref{#1}}

\def\plucker{Pl\"ucker\ }

\def\ie{\emph{i.e.}}


%% file: Main/Metadata/authors.tex

\author{In Cho}
\affiliation{
\institution{Yonsei University}
\country{South Korea}
}

\author{Jeonghwan Cho}
\affiliation{
\institution{Yonsei University}
\country{South Korea}
}

\author{Mijin Yoo}
\affiliation{
\institution{Yonsei University}
\country{South Korea}
}

\author{Gim Hee Lee}
\affiliation{
\institution{National University of Singapore}
\country{Singapore}
}

\author{Seon Joo Kim}
\affiliation{
\institution{Yonsei University}
\country{South Korea}
}



%% file: Main/Sections/abstract.tex
\begin{abstract}
3D Gaussian Splatting (3DGS) achieves high-quality novel-view synthesis by optimizing freely placed primitives in 3D and adaptively densifying them in under-reconstructed regions.
However, this scene-adaptive capacity allocation is largely lost in existing feed-forward 3DGS methods, which commonly regress Gaussians at input pixels and lift them along camera rays.
Such pixel-aligned formulations make the number and placement of primitives depend on image resolution and input viewpoints rather than scene complexity, resulting in dense and often redundant Gaussian sets.
We present ATSplat, a feed-forward 3DGS framework that restores the adaptive allocation capability of 3DGS optimization through Adaptive 3D Tokens.
ATSplat first lifts coarse patch-level depth and camera cues into sparse 3D anchor tokens, forming a compact scaffold of the scene.
Each token is then regressed into local Gaussians with learnable 3D offsets, decoupling primitive placement from input image grids.
An Adaptive Token Expansion module predicts a token-level uncertainty score, supervised by rendering error maps, and selectively expands high-uncertainty tokens through learnable expansion layers.
This sparse-to-adaptive formulation enables ATSplat to concentrate primitives in challenging regions while maintaining a compact representation.
Experiments on two representative datasets, RealEstate10K and DL3DV, show that ATSplat achieves state-of-the-art rendering quality while reducing the number of Gaussians by more than $5.7\times$ compared with dense feed-forward 3DGS methods.
From 12 input images at $512 \times 960$ resolution, ATSplat completes reconstruction in less than a second using a single commercial GPU, and renders high-quality novel views at 1136 FPS ($512 \times 960$) with only 311K Gaussians.
Our project page is at: \url{https://join16.github.io/page-atsplat}
\end{abstract}

%% file: Main/Metadata/ccs.tex
%
%

\ccsdesc[500]{Computing methodologies~Rendering}
\ccsdesc[300]{Computing methodologies~Reconstruction}
\ccsdesc[100]{Computing methodologies~Neural networks}

%% file: Main/Partials/figure_teaser.tex
\begin{teaserfigure}
\includegraphics[width=1\linewidth]{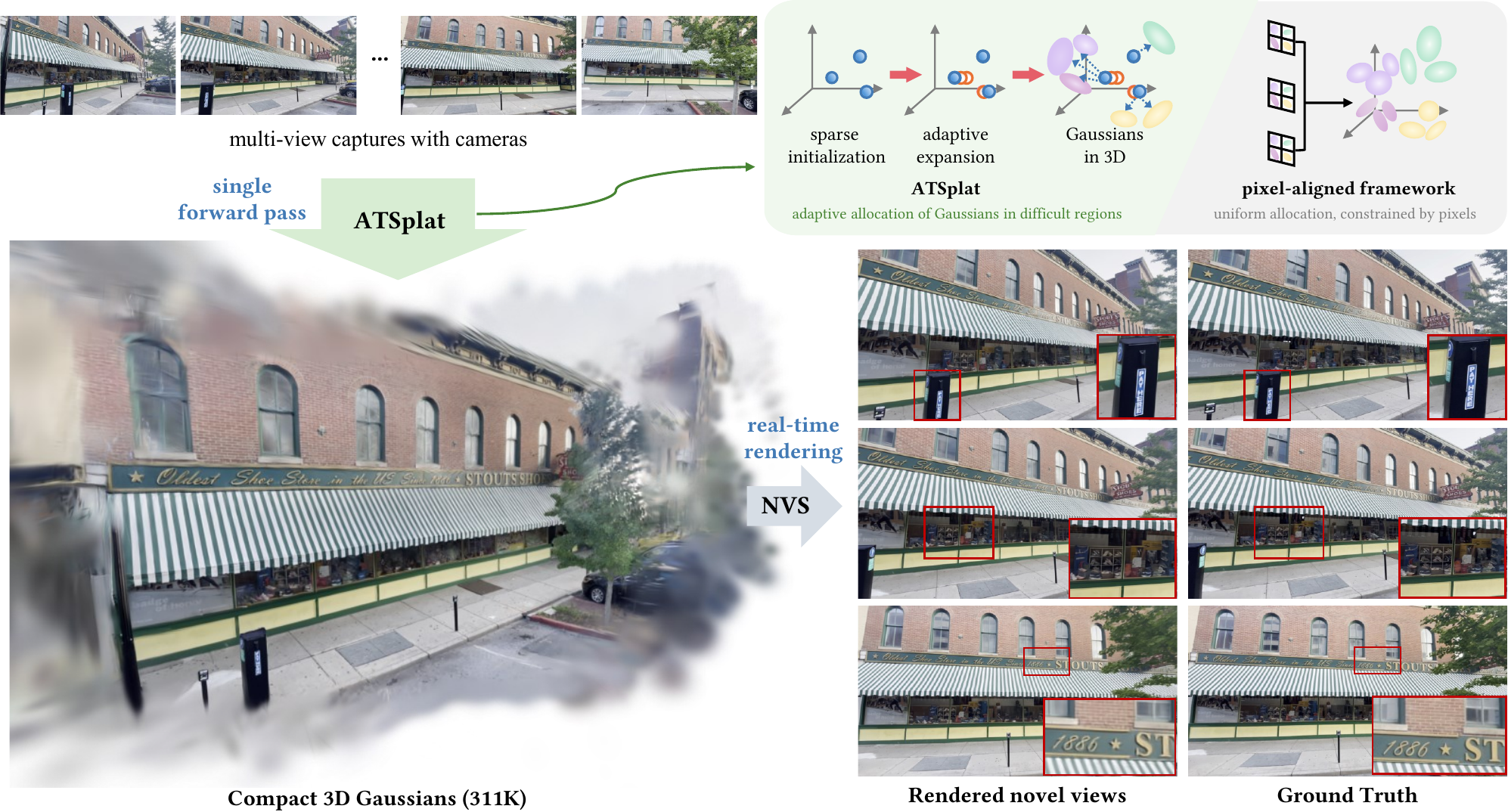}
\caption{
ATSplat reconstructs compact 3D Gaussians from multi-view captures in a single forward pass.
Unlike dense pixel-aligned feed-forward methods that place Gaussians according to input image grids, ATSplat starts from sparse 3D anchor tokens and adaptively expands tokens associated with challenging regions.
This sparse-to-adaptive allocation concentrates representational capacity according to scene complexity, enabling high-quality novel-view rendering with only 311K Gaussians.
With 12 images at $512 \times 960$ resolution, ATSplat completes reconstruction in less than a second, which can be rendered at 1136 FPS.
}
\label{fig:teaser}
\end{teaserfigure}

%% file: Main/Sections/intro.tex
\section{Introduction}

3D Gaussian Splatting (3DGS)~\cite{kerbl20233dgs} has become a powerful representation for novel-view synthesis, achieving high-quality reconstruction with real-time rendering.
The strength of this pipeline lies in the flexibility of its representation: starting from a sparse point cloud as a coarse initialization, primitives are freely positioned throughout 3D space, and adaptive densification redirects representational capacity toward under-reconstructed regions.
This scene-adaptive capacity allocation--primitives placed and refined according to the scene's complexity--established 3DGS as a leading representation for novel-view synthesis.

Yet these advantages are at the cost of per-scene optimization, limiting the practicality of 3DGS in many applications.
A growing body of work therefore seeks to bypass this cost, by leveraging learned priors to reconstruct Gaussian primitives directly from multi-view captures in a single forward pass.
The dominant design in these feed-forward 3DGS frameworks is a pixel-aligned formulation~\cite{charatan2024pixelsplat, chen2024mvsplat, xu2025depthsplat, kang2025ilrm}: one Gaussian is regressed at every input pixel and lifted along its camera ray, reducing primitive placement to depth estimation.
While this formulation simplifies the reconstruction problem and achieves impressive results, it also produces vast amounts of dense, per-pixel Gaussians.
As the number of primitives grows with both image resolution and the number of input views, these methods often suffer from significant rendering inefficiency and storage costs.

More fundamentally, this inefficiency arises from how representation capacity is parameterized and allocated.
By tying Gaussians to input image grids, pixel-aligned frameworks make primitive placement depend on camera sampling rather than reconstruction difficulty.
As a result, they sacrifice a key strength of optimized 3DGS: the ability to allocate capacity adaptively according to scene complexity.
Closing this gap calls for a feed-forward design in which Gaussian placement reflects scene complexity instead of image structure.

We present ATSplat, a feed-forward 3DGS framework that restores scene-adaptive capacity allocation of 3DGS optimization with Adaptive Tokens.
Our framework reinterprets three core design principles of per-scene 3DGS optimization--sparse initialization, free placement in 3D, and adaptive capacity allocation--as feed-forward operations centered around our adaptive anchor tokens.
First, ATSplat lifts coarse patch-level depth and camera information into sparse 3D anchor tokens, forming a compact scaffold of the scene.
Second, each token is decoded into a small set of local Gaussians whose positions are predicted relative to the token anchor, decoupling primitive placement from input image grids.
Third, ATSplat progressively expands tokens associated with challenging regions, concentrating representational capacity where the scene demands it.
Together, this sparse-to-adaptive formulation produces a compact set of Gaussians distributed according to scene complexity.

The central challenge in adaptive allocation is identifying under-reconstructed regions without access to rendering errors, as these errors are not directly observable in a single forward pass.
Our Adaptive Token Expansion module addresses this by learning a per-token uncertainty score, supervised through rendered uncertainty maps to match actual rendering errors.
Coupled with the anchor-based flexible 3D placement, the adaptive expansion enables us to concentrate representational capacity to challenging regions.

Thanks to the flexible and adaptive anchor design, ATSplat uses its representation budget more efficiently compared to dense pixel-aligned feed-forward 3DGS methods.
Experiments on two representative datasets, RealEstate10K~\cite{zhou2018re10k} and DL3DV~\cite{ling2024dl3dv}, demonstrate that ATSplat achieves state-of-the-art rendering quality while reducing the number of Gaussians by more than $5.7\times$.
These results suggest that high-quality feed-forward 3DGS arises from how representation capacity is allocated, not from how densely it is sampled. By restoring core design principles of 3DGS, ATSplat achieves state-of-the-art quality with more compact representation and real-time rendering.

Our key contributions can be summarized as follows:
\begin{itemize}
    \item We propose ATSplat, a feed-forward 3DGS framework built around adaptive 3D anchor tokens that restores the scene-adaptive capacity allocation of 3DGS in feed-forward setups.
    \item We introduce an Adaptive Token Expansion module that identifies and progressively expands tokens in challenging regions without directly accessing rendering errors.
    \item ATSplat achieves state-of-the-art rendering quality while using over $5.7\times$ fewer Gaussians than previous dense pixel-aligned methods, demonstrating the importance of scene-adaptive capacity allocation for compact feed-forward 3DGS.
\end{itemize}

%% file: Main/Sections/related_work.tex
\section{Related Work}
\subsection{3D Representations and Novel-View Synthesis}
Neural radiance fields (NeRF) represent a scene as a continuous radiance-density field and synthesize novel views through volumetric rendering~\cite{mildenhall2020nerf}.
A large body of work has improved NeRF in several aspects, which include anti-aliasing~\cite{barron2021mipnerf}, training speed~\cite{muller2022instant}, explicit factorization~\cite{chen2022tensorf, chan2022eg3d}, and generalizable radiance fields~\cite{yu2021pixelnerf, wang2021ibrnet, chen2021mvsnerf}.

Despite their high fidelity, many neural-field methods require dense ray sampling and/or costly per-scene optimization. 
3D Gaussian Splatting (3DGS)~\cite{kerbl20233dgs} instead represents scenes with a set of anisotropic Gaussian primitives and optimizes them via differentiable rasterization, enabling high-quality real-time rendering.
A rapidly growing body of follow-up work has extended it along many directions, including surface reconstruction~\cite{huang20242dgs, guedon2024sugar, yu2024gof}, anti-aliased rendering~\cite{yu2024mipsplatting}, dynamic scene modeling~\cite{wu20244dgaussians, yang20244dgs}, and compact, structured representations~\cite{lu2024scaffold}.
A key strength of optimization-based 3DGS lies in adaptive primitive allocation based on scene complexity, which is enabled by 3D placement of Gaussians and adaptive density control.

\subsection{Feed-forward 3D Gaussian Splatting}
Instead of optimizing Gaussians per scene, recent feed-forward 3DGS methods directly predict Gaussian primitives from multi-view captures, thereby addressing expensive per-scene optimization costs.
Splatter Image~\cite{szymanowicz2024splatter} demonstrates single-view object reconstruction with Gaussian maps, while pixelSplat predicts 3D Gaussians from posed image pairs by lifting pixel-aligned predictions along camera rays~\cite{charatan2024pixelsplat}.
MVSplat~\cite{chen2024mvsplat} further improves this feed-forward framework with a plane-sweep cost volume, and DepthSplat~\cite{xu2025depthsplat} further connects multi-view depth estimation with Gaussian prediction using depth priors.
Subsequent works extend feed-forward 3DGS frameworks to 360-degree~\cite{chen2024mvsplat360}, unposed~\cite{hong2024pf3plat}, and unconstrained settings~\cite{jiang2025anysplat}.
Large reconstruction models (LRM) further scale this pixel-aligned paradigm with large transformer backbones~\cite{xu2024grm, tang2024lgm, zhang2024gslrm}.
More recently, iLRM~\cite{kang2025ilrm} decouples the viewpoint tokens from input images and predicts Gaussians in a separate, low-resolution grid, yielding a more compact set of Gaussians.
However, such uniform spatial downsampling reduces primitives across the entire scene regardless of scene complexity.
This leads to under-allocated Gaussians in complicated regions and degrades rendering quality in these regions.

Although these methods achieve impressive reconstruction speed and rendering quality, most of them follow dense pixel-aligned formulations: the number of primitives are strongly tied to the image resolution and the number of views.
As a result, simple and textureless regions can produce many redundant Gaussians, while challenging regions cannot receive additional capacity that they demand.
ATSplat overcomes this limitation through sparse-to-adaptive capacity allocation, where our adaptive anchor tokens are selectively expanded based on reconstruction difficulty.
Since this expansion proceeds inside the decoder, the expanded tokens are further refined by cross-attending to input images, allocating not only primitives but also computation to challenging regions.

\subsection{Compact Feed-forward 3D Gaussian Splatting}
Several recent works attempt to address the limitation of dense pixel-aligned Gaussian predictions.
FreeSplat and FreeSplat++ fuse overlapping pixel-aligned Gaussians across views and remove redundant or floating primitives for indoor 3D scene reconstruction~\cite{wang2024freesplat, wang2025freesplatpp}.
Gaussian Graph Network models cross-view relations among Gaussian groups and pools them into a more efficient representation~\cite{zhang2024ggn}, while Fuse-and-Refine aggregates pixel-aligned primitives into a canonical 3D space before refinement~\cite{wang2026fuseandrefine}. 
Generative Densification~\cite{nam2025genden} takes a complementary direction: rather than reducing primitives, it learns a feed-forward densification module that upsamples features to generate fine Gaussians for high-frequency details.
These methods mostly operate on top of reconstructed pixel-aligned Gaussians, and introduce post-hoc processing stages.

Concurrent to our work, an emerging line of works move more directly beyond pixel-aligned formulations.
VolSplat~\cite{wang2025volsplat} replaces image-grid alignment with voxel-aligned prediction, and SparseSplat~\cite{zhang2026sparsesplat} adaptively samples Gaussians according to local information richness.
Several works adaptively sample Gaussians from the image grid, with multi-granularity Gaussians~\cite{kim2026f4splat} or adaptive locations within image grids~\cite{moreau2025offhegrid}.
Other concurrent approaches decouple Gaussian prediction from pixels using 3D anchored feature volumes~\cite{zhang2026anchorsplat}, or global learnable tokens~\cite{an2025c3g, ren2026tokengs, itkin2026globalsplat}.

ATSplat is aligned with this emerging direction, but differs in how adaptive capacity is allocated. 
Instead of starting from a dense pixel-aligned Gaussian set, a fixed voxel lattice, or purely learnable global tokens, ATSplat constructs a scene-conditioned scaffold of sparse 3D anchor tokens from coarse depth and cameras.
Each anchor is regressed into a set of local Gaussians with learnable 3D offsets, allowing primitives to be freely placed beyond camera rays.
The Adaptive Token Expansion module identifies under-reconstructed tokens and selectively expands them.
This sparse-to-adaptive formulation aims to fill the missing component in feed-forward 3DGS: allocating representation and computation capacity according to reconstruction difficulty, rather than input image structure.

%% file: Main/Sections/method.tex
\input{Main/Partials/figure_method}
\section{Problem Formulation}
We aim to reconstruct a 3D Gaussian representation of a scene from posed multi-view images in a single forward pass, avoiding the costly per-scene optimization required by standard 3DGS.
Given $V$ multi-view images with camera poses $\{\mathbf{I}_v, \boldsymbol{\pi}_v\}_{v=1}^{V}$, a feed-forward 3DGS model predicts a set of Gaussian attributes:
\begin{equation}
    f_\theta:\;
    \{\mathbf{I}_v, \boldsymbol{\pi}_v\}_{v=1}^{V}
    \;\longmapsto\;
    \{(\boldsymbol{\mu}_g, \mathbf{q}_g, \mathbf{s}_g, \alpha_g, \mathbf{SH}_g)\}_{g=1}^{G},
\end{equation}
where each Gaussian is parameterized by its center $\boldsymbol{\mu}_g$, rotation quaternion $\mathbf{q}_g$, scale $\mathbf{s}_g$, opacity $\alpha_g$, and spherical harmonics $\mathbf{SH}_g$.

Many feed-forward approaches regress Gaussians in a pixel-aligned manner, where each primitive center is determined by a predicted depth and the corresponding camera ray:
\begin{equation}
    \boldsymbol{\mu}_v(\mathbf{x}) \;=\; \mathbf{o}_v + d_v(\mathbf{x})\,\mathbf{r}_v(\mathbf{x}),
\end{equation}
where $\mathbf{o}_v\in \mathbb{R}^3$ and $\mathbf{r}_v(\mathbf{x}) \in \mathbb{R}^3$ are the camera origin and ray direction at pixel $\mathbf{x}$.
This simplifies center prediction to per-pixel depth estimation, but ties each primitive to an input pixel: the total Gaussian budget and its spatial distribution are determined by input pixel grids rather than the underlying scene.
As a result, the pixel-aligned formulation produces many redundant primitives in trivial regions while under-allocating capacity to challenging regions.
We therefore treat this inefficiency as a formulation-level problem: a feed-forward 3DGS model should allocate representations according to the scene's reconstruction difficulty rather than input pixel grids.

\section{ATSplat Framework}
We introduce ATSplat, a feed-forward 3DGS framework that restores the scene-adaptive Gaussian placement of per-scene 3DGS optimization.
Our key idea is to recast three guiding principles of 3DGS--sparse initialization, free 3D placement, adaptive capacity allocation--as feed-forward operations centered on a set of 3D anchor tokens.
Concretely, sparse initialization is achieved by unprojecting coarse patches into 3D, free 3D placement by regressing local primitive offsets from each anchor, and adaptive capacity allocation by expanding high-uncertainty anchors during decoding.

\fref{fig:method} (a) illustrates an overview of the framework.
Given multi-view images, a multi-view encoder first extracts coarse patch features and predicts patch-level depths.
The depths are unprojected with the camera rays, assigning coarse patch features to sparse 3D locations that form an anchor scaffold.
An image-to-3D decoder then refines these anchor tokens by cross-attending to fine-grained image features.
Within the decoder, an Adaptive Token Expansion (ATE) module selectively expands tokens associated with under-reconstructed regions, increasing representational capacity only where needed.
Finally, each refined token is decoded into a set of local Gaussians whose centers are predicted as 3D offsets from its anchor, decoupling primitive placement from the input pixel grid.
The following paragraphs describe details of each component.

\paragraph{Multi-view image encoder.}
The encoder extracts cross-view patch features from the input images at the coarse resolution.
Each view is first tokenized into coarse patches using a frozen DINO backbone.
We also add \plucker raymap embeddings to the patch tokens to inject camera geometry.
A multi-view transformer then flattens the patch tokens across all views and applies global self-attention to produce cross-view image features.

\paragraph{Initializing sparse anchor tokens.}
We then initialize a sparse 3D scaffold by computing a 3D coordinate for each encoded patch, which serve as initial anchor tokens.
For each patch feature $\mathbf{f}_i \in \mathbb{R}^{C}$ at pixel $\mathbf{x}_i$ in view $v_i$, 
we predict a patch-level depth $\hat{d}_i$ from $\mathbf{f}_i$ with a lightweight MLP and unproject it along the corresponding ray:
\begin{equation}
    \mathbf{p}_i \;=\; \mathbf{o}_{v_i} + \hat{d}_i \, \mathbf{r}_{v_i}(\mathbf{x}_i).
\end{equation}
The pair $(\mathbf{p}_i, \mathbf{f}_i)$ defines a 3D anchor token--anchored at $\mathbf{p}_i$ and carrying the encoded feature $\mathbf{f}_i$.
We further inject local 3D context by aggregating each anchor with its $k$-nearest neighbors (kNN) via a PointNet-style operator~\cite{qi2017pointnet}.

\paragraph{Image-to-3D decoder.}
The decoder refines anchor tokens by injecting fine-grained image information through cross-attention.
For each view, we extract fine patch features at twice the coarse resolution using a lightweight per-view feature extractor.
This adds \plucker raymap embeddings to the patches and applies two per-view self-attention layers.
The decoder then applies a stack of $L$ blocks, each composed of an ATE module and cross-attention layers.
The ATE module selectively expands anchors associated with under-reconstructed regions, progressively increasing representational capacity where needed. We describe details of ATE in \sref{sec:method_ate}.

\paragraph{Anchors to local 3D Gaussians.}
Each refined anchor token $\hat{f}_i$ is finally regressed into a set of $K$ local Gaussian primitives.
A lightweight Gaussian head, consists of 2-layer MLPs, maps the anchor feature to $K$ sets of Gaussian attributes:
\begin{equation}
    \bigl\{(\Delta \boldsymbol{\mu}_{i,k},\, \mathbf{q}_{i,k},\, \mathbf{s}_{i,k},\, 
    \alpha_{i,k},\, \mathbf{SH}_{i,k})\bigr\}_{k=1}^{K} 
    \;=\; \mathrm{MLP}(\mathbf{\hat{f}}_i),
\end{equation}
and each Gaussian center is placed relative to the anchor as 
\begin{equation}
  \boldsymbol{\mu}_{i,k} = \mathbf{p}_i + \Delta \boldsymbol{\mu}_{i,k},
\end{equation}
which places primitive positions freely from the input pixel grid.

\section{Adaptive Token Expansion}
\label{sec:method_ate}
Unlike per-scene optimization, a feed-forward model cannot directly access rendering errors or gradients during decoding, making it non-trivial to identify under-reconstructed regions.
Therefore, we approximate the rendering error associated with each anchor through a lightweight MLP, 
and use these predictions to selectively expand anchors that require additional representational capacity.

At each decoder block, we estimate an uncertainty score for every anchor token $\mathbf{f}_i$ through a lightweight MLP $g_\phi$:
\begin{equation}
    u_i \;=\; g_\phi(\mathbf{f}_i) \;\in\; \mathbb{R}.
\end{equation}
We select top fraction $\rho_l$ of anchors with the highest scores as expansion targets.
Each selected feature is passed through a linear projection that produces $M$ residual features, yielding $M$ child tokens that share the parent anchor coordinate $\mathbf{p}_i$:
\begin{equation}
    [\Delta\mathbf{f}_{i,1},\,\dots,\,\Delta\mathbf{f}_{i,M}] \;=\; \mathbf{W}\,\mathbf{f}_i,
    \qquad
    \mathbf{f}_{i,m} \;=\; \mathbf{f}_i + \Delta\mathbf{f}_{i,m}.
\end{equation}
The expanded tokens together with the unselected tokens are concatenated and passed to the next decoder block.
This design identifies under-reconstructed regions in a single forward pass, without auxiliary renderings or back-propagation at inference.

\paragraph{Learning uncertainty in 2D space.}
To align the predicted uncertainty with the actual reconstruction difficulty, we supervise $g_\phi^{(l)}$ using 2D reconstruction errors.
After the $l$-th decoder block, we apply the Gaussian head to the current anchors to obtain an intermediate Gaussian set $\mathcal{G}^{(l)}$.
For each anchor $i$, the score $u_i^{(l)}$ is attached as a scalar attribute to its $K$ Gaussians and rasterized via standard alpha-compositing, producing a 2D uncertainty map $\hat{\mathbf{U}}^{(l)}$.
We supervise $\hat{\mathbf{U}}^{(l)}$ against the error map $\mathbf{U}^{(l)}$ of $\mathcal{G}^{(l)}$:
\begin{equation}
    \mathcal{L}_{\mathrm{unc}}^{(l)}
    \;=\;
    \bigl\|\, \hat{\mathbf{U}}^{(l)} - \mathrm{sg}\bigl(\mathbf{U}^{(l)}\bigr) \,\bigr\|_1,
\end{equation}
where $\mathrm{sg}(\cdot)$ stops gradients into the Gaussian parameters so that this loss updates only $g_\phi^{(l)}$.
The error map $\mathbf{U}^{(l)}$ is computed using the rendered results $\mathcal{G}^{(l)}$ and ground-truth images.
We empirically use D-SSIM to compute these intermediate error maps.

\section{Training}
We train ATSplat end-to-end with a final rendering loss, intermediate rendering losses, and uncertainty supervision.
The final rendering loss combines MSE with a perceptual term:
\begin{equation}
    \mathcal{L}_{\mathrm{render}}
    \;=\; \mathcal{L}_{\mathrm{MSE}} + \lambda_{\mathrm{p}} \cdot \mathcal{L}_{\mathrm{perceptual}},
\end{equation}
where we use $\lambda_p = 0.5$.
To train the uncertainty heads at intermediate decoder blocks, we additionally supervise each intermediate Gaussian set $\mathcal{G}^{(l)}$ with an auxiliary rendering loss $\mathcal{L}_{\mathrm{interm}}^{(l)}$.
For efficiency, we replace the perceptual term with $\mathcal{L}_{\mathrm{D\text{-}SSIM}}$ in the intermediate rendering loss.
The full objective is
\begin{equation}
    \mathcal{L}
    \;=\; \mathcal{L}_{\mathrm{render}}
    \;+\; \sum_{l=1}^{L} \bigl(
        \lambda_{\mathrm{int}} \cdot \mathcal{L}_{\mathrm{interm}}^{(l)}
        \;+\; \lambda_{\mathrm{unc}} \cdot \mathcal{L}_{\mathrm{unc}}^{(l)},
    \bigr).
\end{equation}
where we use $\lambda_{\mathrm{interm}}=0.5$ and $\lambda_{\mathrm{unc}}=0.1$ in all experiments.

%% file: Main/Partials/figure_method.tex
\begin{figure*}[t]
\centering
\includegraphics[width=1\linewidth]{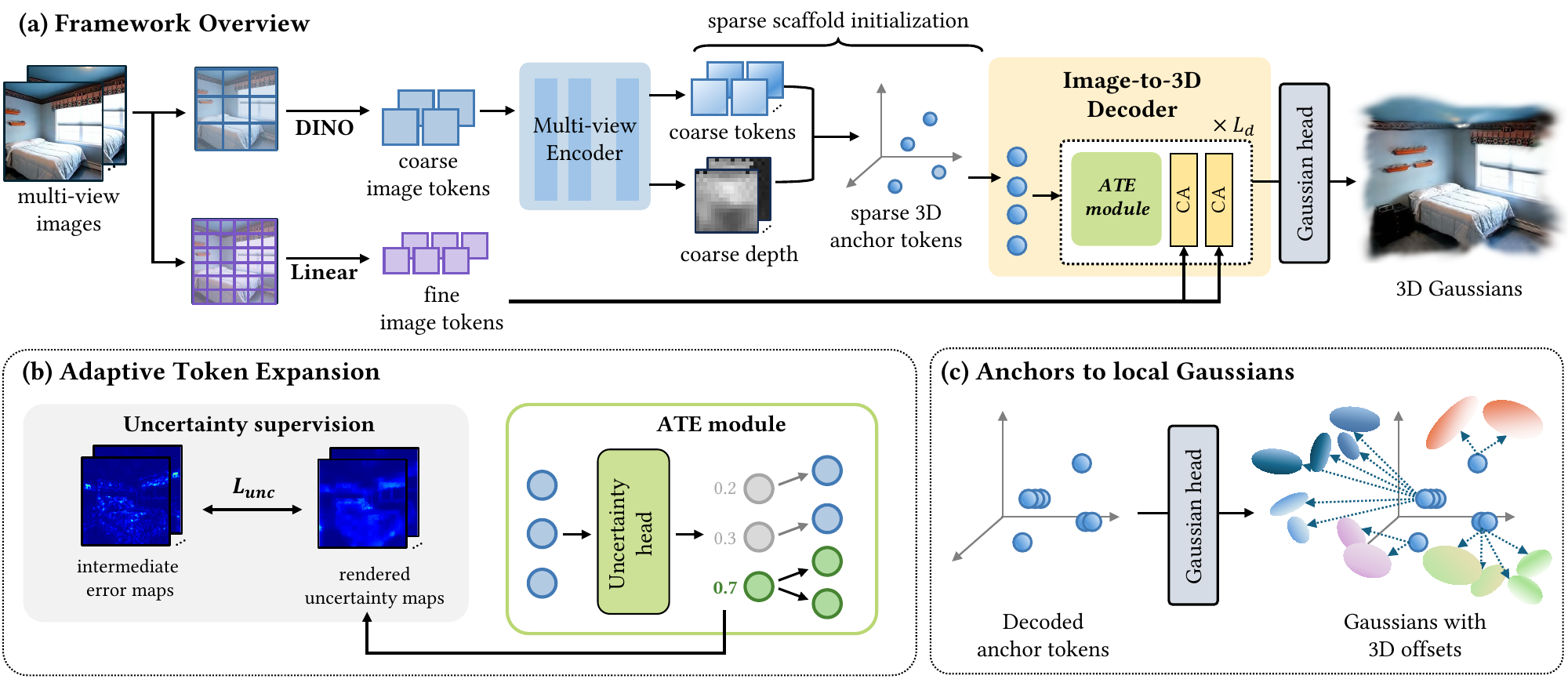}
\caption{
\textbf{(a) Framework overview.} 
Given posed multi-view images, a multi-view image encoder extracts coarse patch features and estimates patch-level depths.
The predicted depths and patch features are used to initialize a sparse set of 3D anchor tokens, serving as a sparse scene scaffold.
An image-to-3D decoder refines these anchors through $L_d$ decoder blocks, with an Adaptive Token Expansion (ATE) module that allocates additional capacity to under-reconstructed regions. 
A Gaussian head then decodes each refined anchor into a set of local 3D Gaussians.
\textbf{(b) Adaptive Token Expansion.} 
At each decoder block, a lightweight MLP predicts a per-anchor uncertainty score, supervised by 2D error maps computed from intermediate Gaussians. Anchors with high uncertainty scores are expanded into multiple tokens, increasing representational capacity where needed.
\textbf{(c) Anchors to local Gaussians.}
Each refined anchor token is regressed into K local Gaussians whose centers are placed as 3D offsets relative to the anchor, decoupling primitive positions from the input pixel grid.
}
\label{fig:method}
\end{figure*}

%% file: Main/Sections/experiment.tex
\section{Experiments}

\subsection{Experimental Setup}
\paragraph{Datasets.}
We train and evaluate ATSplat on two widely used datasets for feed-forward novel-view synthesis: RealEstate10K \cite{zhou2018re10k} and DL3DV~\cite{ling2024dl3dv}.
RealEstate10K consists of home-tour videos collected from YouTube and mainly contains bounded indoor scenes.
DL3DV provides larger-scale captures of both indoor and outdoor environments, covering larger spatial extents, more diverse scene layouts and complex geometry.
Following previous works~\cite{charatan2024pixelsplat, chen2024mvsplat, xu2025depthsplat}, we use the same preprocessing steps and train/test splits.

\paragraph{Evaluation protocols.}
We report PSNR, SSIM, and LPIPS~\cite{zhang2018lpips} for rendering quality, 
and measure efficiency by the number of Gaussians and inference runtime.
Runtime is measured on a single RTX 3090 GPU unless otherwise specified.
For RealEstate10K, we use the evaluation viewpoint indices provided in pixelSplat \cite{charatan2024pixelsplat}.
For DL3DV, we follow the protocol of DepthSplat~\cite{xu2025depthsplat} and evaluate under 2, 4, 6 input-view setups.
We additionally evaluate results under varying viewpoint overlaps following NoPoSplat~\cite{ye2025noposplat} in the supplementary.

\paragraph{Baselines.}
We compare ATSplat with recent feed-forward 3DGS methods, including pixelSplat~\cite{charatan2024pixelsplat}, MVSplat~\cite{chen2024mvsplat}, DepthSplat~\cite{xu2025depthsplat}, NoPoSplat~\cite{ye2025noposplat}, and iLRM~\cite{kang2025ilrm}, where applicable.
For quantitative comparisons, we use reported numbers when methods follow the same evaluation protocol, and otherwise evaluate available public checkpoints under the same evaluation setup.
For qualitative comparisons on DL3DV, we compare primarily with DepthSplat, as it is the only feed-forward 3DGS baseline with a publicly available checkpoint under this setting.
For high-resolution evaluations, we further compare with 3DGS~\cite{kerbl20233dgs} and Mip-Splatting~\cite{yu2024mipsplatting} as optimization-based baselines. These methods are initialized with sparse SfM~\cite{schonberger2016colmap} points of the input views and optimized for 30K iterations.

\input{Main/Partials/table_re10k}
\input{Main/Partials/table_dl3dv}
\paragraph{Implementation details.}
ATSplat consists of a multi-view encoder with 12 global self-attention layers, followed by a decoder with 4 blocks.
We use a frozen DINOv2-B~\cite{oquab2023dinov2} model with a 768 dimension for the coarse patch branch.
An ATE module is employed in all decoder blocks except the first one, with token selection ratios $\rho_l$ of $0.5$, $0.5$, and $0.25$, respectively, and an expansion ratio of $2$.
Each anchor token is finally decoded into $K=16$ local Gaussians.
ATSplat is first trained on RealEstate10K with 2 input views using 4 RTX 4090 GPUs for about two days, serving as the base model.
It is further trained on DL3DV with 6 input views using 4 H200 GPUs for two days.
For the high-resolution setting, we further fine-tune this model with 10 input views using 8 H200 GPUs for less than two days.
Please refer to the supplementary for more details.

\input{Main/Partials/table_dl3dv_hr}
\subsection{Results}
We first evaluate ATSplat on common benchmark setups adopted in many previous works~\cite{charatan2024pixelsplat, chen2024mvsplat, xu2025depthsplat, kang2025ilrm}. These include comparisons on RealEstate10K with two input views at $256\times 256$ resolution, and comparisons on DL3DV at $256\times 448$ resolution with various input view settings.
We further encourage the readers to check the supplementary for more qualitative results and video comparisons.

\paragraph{RealEstate10K}
\tref{tab:re10k} reports two-view results on RealEstate10K.
ATSplat achieves rendering quality comparable to state-of-the-art feed-forward methods while using only $23$K Gaussians, corresponding to a $5.7\times$ reduction compared to dense pixel-aligned formulations.
\fref{fig:qualitative_re10k} further confirms this, showing accurate reconstruction of fine details and geometry.
By allocating primitives from sparse anchors and selectively expanding them, ATSplat maintains high-quality results with a much smaller representation compared to dense pixel-aligned baselines.
We further provide evaluations with varying viewpoint overlaps in the supplementary, which highlights the efficacy of our 3D anchor designs.

\paragraph{DL3DV}
We next evaluate ATSplat on DL3DV, which contains more challenging scenes with complex geometry.
\tref{tab:quantitative_dl3dv} reports results at $256\times448$ resolution using 2, 4, and 6 input views.
Across all input-view settings, ATSplat outperforms prior feed-forward baselines while using substantially fewer Gaussians, demonstrating that the sparse-to-adaptive formulation remains effective in larger and more diverse scenes.
\fref{fig:qualitative_dl3dv} also supports this observation, with thin structures and intricate geometry faithfully recovered.
In particular, iLRM~\cite{kang2025ilrm} reduces the number of Gaussians by predicting them at a lower spatial resolution.
However, this uniform reduction decreases Gaussian density across the entire scene, under-allocating capacity to regions with fine structures or intricate geometry.
In contrast, ATSplat begins with sparse anchors and selectively expands them in uncertain regions, achieving high-quality results with a compact Gaussian set.

\input{Main/Partials/figure_child_behaviors}
\subsection{High-Resolution Novel-View Synthesis}
We further evaluate ATSplat on DL3DV at $512\times960$ resolution to demonstrate that our sparse-to-adaptive design scales to high-resolution rendering.
For efficiency, we downsample the images used by the coarse patch branch (\ie, DINO) by $2\times$.
We also increase the selection ratio $\rho_1$ of the first ATE block from $0.5$ to $0.8$, as the finer details exposed at this resolution benefit from more expansion.
We compare with available feed-forward 3DGS methods as well as optimization-based 3DGS baselines.
For DepthSplat~\cite{xu2025depthsplat}, we use the public checkpoint trained at $448\times768$ resolution, the only configuration with released weights applicable to this setup.

As shown in \tref{tab:dl3dv_hr}, ATSplat reconstructs high-resolution scenes in a single forward pass and completes inference in less than one second.
It achieves compelling rendering results while using far fewer Gaussians than previous dense pixel-aligned feed-forward methods, which can also be found in \fref{fig:qualitative_dl3dv_960p}.
This efficiency becomes important at high resolution, where pixel-aligned methods typically produces more than a million Gaussians.

\subsection{Analysis}
To better understand the behaviors of our core components, we provide analyses on expanded tokens and uncertainty predictions.

\paragraph{Behavior of expanded tokens.}
We first analyze how ATE increases local representation capacity.
For each expanded token, we compute a representative 3D location by averaging the centers of the Gaussians decoded from that token.
Although ATE does not explicitly update anchor coordinates during expansion, the expanded token features produce distinct local Gaussian offsets and features.
As shown in \fref{fig:child_behaviors}, child tokens spread around their parent tokens and cover different nearby regions instead of collapsing to the same position.
This supports that ATE learns to refine local geometry and appearance based on the detailed scene content.

\input{Main/Partials/table_ablation_anchor}
\input{Main/Partials/figure_uncertainty_vis}
\paragraph{Predicted uncertainty and selected tokens.}
We further analyze whether the learned uncertainty provides a meaningful proxy for reconstruction difficulty.
As shown in \fref{fig:uncertainty_vis}, the predicted uncertainty is high around regions with larger rendering errors, including fine structures and visually complex areas. The selected tokens are concentrated in these uncertain regions, indicating that ATE successfully identifies where additional Gaussians are needed and places more capacity to these regions with diversified expanded tokens.



\subsection{Ablations}
We examine the effects of our two core components through the ablations: 3D anchor designs and uncertainty-guided selection.
All ablations are conducted on RealEstate10K with two input views.

\paragraph{3D anchor designs.}
We compare our anchor-offset formulation with three alternatives.
The first is a pixel-aligned ray-depth formulation, where Gaussians are tied to input pixels and lifted along camera rays.
The second uses learnable initial tokens without coarse scene-specific geometry.
The third directly regresses absolute 3D Gaussian centers instead of local offsets around coarse anchors.

As shown in \tref{tab:ablation_anchor}, the proposed anchor-offset design achieves the best reconstruction quality.
The pixel-aligned variant constrains Gaussians to input pixels, which limits the flexibility of our adaptive expansion.
The latter two variants lack scene-specific geometric priors, leaving Gaussian placement to be learned without spatial guidance.
These results indicate that coarse 3D anchors provide meaningful scene structure, while local offsets preserve the flexibility needed for accurate Gaussian placement.

\input{Main/Partials/table_ablation_ate}
\paragraph{ATE module.}
We further ablate the selection strategy and expansion in ATE.
All selection-based variants use the same expansion ratio and produce the same number of Gaussians.
As shown in \tref{tab:ablation_ate}, random selection and FPS provide limited gains as they allocate capacity without targeting difficult
regions.
We also compare with a straight-through estimator (STE)~\cite{ste}, following a prior upsampling design~\cite{nam2025genden}, but find that it does not improve over simple heuristic selection.
Removing token expansion entirely leads to a notable performance drop.
Our full design achieves the best performance, demonstrating the effectiveness of both uncertainty-guided selection and learnable expansion.

%% file: Main/Partials/table_re10k.tex
\begin{table}[t]
\setlength{\tabcolsep}{3pt}
\centering
\caption{
\textbf{Quantitative results on RealEstate10K at $256\times 256$ resolution} with 2 input views. Inference times for GS-LRM and LongLRM are omitted, due to the absence of public codes for this setting.
}
\label{tab:re10k}
\footnotesize
\begin{tabular}{lccccc}
\toprule
Method & PSNR$\uparrow$ & SSIM$\uparrow$ & LPIPS$\downarrow$ & \#Gauss.$\downarrow$ & Time(s)$\downarrow$ \\
\midrule
PixelSplat \cite{charatan2024pixelsplat} & 25.89 & 0.858 & 0.142 & 131K & 0.127 \\
MVSplat \cite{chen2024mvsplat} & 26.39 & 0.869 & 0.128 & 131K & 0.044 \\
GS-LRM \cite{zhang2024gslrm} & 28.10 & 0.892 & 0.114 & 131K & - \\
DepthSplat \cite{xu2025depthsplat} & 27.47 & 0.889 & 0.114 & 131K & 0.067 \\
LongLRM \cite{ziwen2025long} & \underline{28.54} & 0.895 & \textbf{0.109} & 131K & - \\
iLRM \cite{kang2025ilrm} & \textbf{28.65} & \underline{0.900} & \underline{0.110} & 131K & 0.025 \\ 
TokenGS \cite{ren2026tokengs} & 28.41 & \textbf{0.903} & 0.135 & 262K & 0.066 \\
Ours & 28.46 & 0.894 & 0.118 & \textbf{23K} & \textbf{0.022} \\
\bottomrule
\end{tabular}
\end{table}

%% file: Main/Partials/table_dl3dv.tex
\begin{table*}[t!]
\setlength{\tabcolsep}{6pt}
\centering
\caption{
\textbf{Quantitative comparisons on DL3DV at $256 \times 448$ resolution.}
*: reported only at 6 views, as both code and pretrained weights are not publicly available for the other setups.
iLRM predicts pixel-aligned Gaussians at $2\times$ downsampled resolution, which results in $4\times$ fewer Gaussians.
}
\label{tab:quantitative_dl3dv}
\footnotesize
\begin{tabular}{l cccc cccc cccc}
\toprule
& \multicolumn{4}{c}{2 input views} & \multicolumn{4}{c}{4 input views} & \multicolumn{4}{c}{6 input views} \\
\cmidrule(lr){2-5} \cmidrule(lr){6-9} \cmidrule(lr){10-13}
Method
 & PSNR$\uparrow$ & SSIM$\uparrow$ & LPIPS$\downarrow$ & \#Gauss.$\downarrow$
 & PSNR$\uparrow$ & SSIM$\uparrow$ & LPIPS$\downarrow$ & \#Gauss.$\downarrow$
 & PSNR$\uparrow$ & SSIM$\uparrow$ & LPIPS$\downarrow$ & \#Gauss.$\downarrow$ \\
\midrule
MVSplat \cite{chen2024mvsplat}
 & 17.54 & 0.529 & 0.402 & 229K
 & 21.63 & 0.721 & 0.233 & 458K
 & 22.93 & 0.775 & 0.193 & 688K \\
DepthSplat \cite{xu2025depthsplat}
 & 19.31 & 0.615 & \textbf{0.310} & 229K
 & 23.12 & 0.780 & 0.178 & 458K
 & 24.19 & 0.823 & 0.147 & 688K \\
iLRM* \cite{kang2025ilrm}
 & -- & -- & -- & --
 & -- & -- & -- & --
 & 25.60 & 0.830 & 0.168 & 172K \\
TokenGS \cite{ren2026tokengs}
 & 19.35 & 0.609 & 0.440 & 262K
 & 23.02 & 0.747 & 0.326 & 262K
 & 23.69 & 0.766 & 0.311 & 262K \\
SparseSplat* \cite{zhang2026sparsesplat}
 & -- & -- & -- & --
 & -- & -- & -- & --
 & 24.20 & 0.817 & 0.168 & 150K \\
Ours
 & \textbf{20.07} & \textbf{0.630} & \underline{0.332} & \textbf{40K}
 & \textbf{25.67} & \textbf{0.827} & \textbf{0.172} & \textbf{80K}
 & \textbf{27.28} & \textbf{0.868} & \textbf{0.138} & \textbf{120K} \\
\bottomrule
\end{tabular}
\end{table*}

%% file: Main/Partials/table_dl3dv_hr.tex
\begin{table}[t]
\setlength{\tabcolsep}{3pt}
\centering
\footnotesize
\caption{
\textbf{Quantitative comparisons on DL3DV at $512\times 960$ resolution} with 100-frame baselines.
The top group lists optimization-based methods, and the bottom group lists feed-forward 3DGS methods.
*: Runtime of DepthSplat is measured using a single A6000 GPU due to out-of-memory.
}
\label{tab:dl3dv_hr}
\begin{tabular}{lccccc}
\toprule
Method & PSNR$\uparrow$ & SSIM$\uparrow$ & LPIPS$\downarrow$ & \#Gauss.$\downarrow$ & Time(s)$\downarrow$ \\
\midrule
3DGS \cite{kerbl20233dgs}               & 22.87 & 0.758 & \textbf{0.229} & \underline{553K} & > 10 m \\
Mip-Splatting \cite{yu2024mipsplatting} & 22.52 & 0.746 & \underline{0.240} & 676K & > 10 m \\
\midrule
DepthSplat* \cite{xu2025depthsplat}     & 21.33 & 0.740 & 0.270 & 5898K & 0.758 s \\
iLRM \cite{kang2025ilrm}                & \underline{24.35} &\underline{ 0.781} & 0.256 & 1474K & \underline{0.700 s} \\
Ours                                    & \textbf{24.85} & \textbf{0.794} & 0.241 & \textbf{311K} & \textbf{0.677 s} \\
\bottomrule
\end{tabular}
\end{table}

%% file: Main/Partials/figure_child_behaviors.tex
\begin{figure}[t]
\centering
\includegraphics[width=1\linewidth]{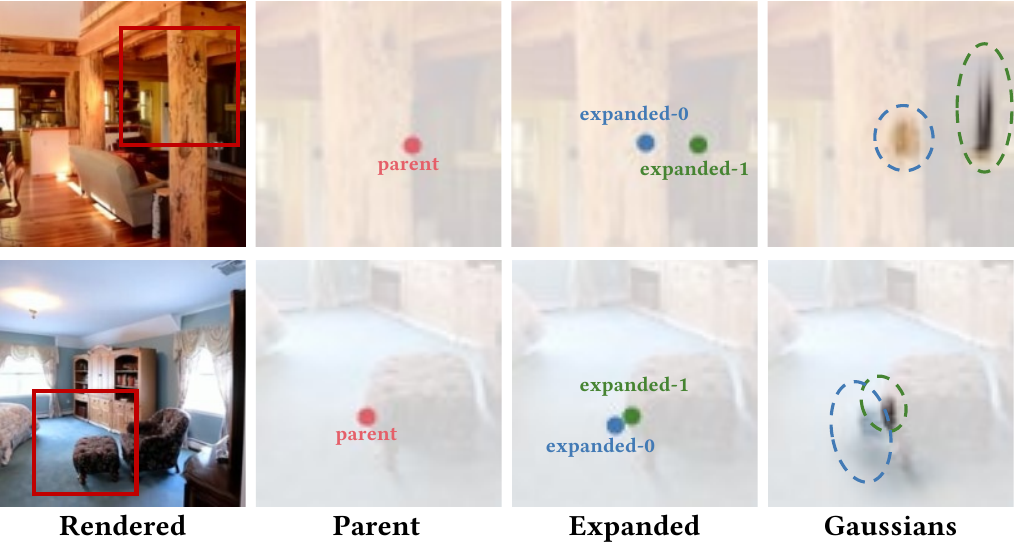}
\caption{
\textbf{Visualization of expanded tokens.}
The coordinate of each expanded token is computed as the mean center of its decoded Gaussians.
}
\label{fig:child_behaviors}
\end{figure}

%% file: Main/Partials/table_ablation_anchor.tex
\begin{table}[t]
\setlength{\tabcolsep}{7pt}
\centering
\small
\caption{\textbf{Ablation results on the 3D anchor designs.}}
\label{tab:ablation_anchor}
\begin{tabular}{llccc}
\toprule
Anchor init. & Gaussian center & PSNR$\uparrow$ & SSIM$\uparrow$ & LPIPS$\downarrow$ \\
\midrule
None         & Ray $+$ depth & 27.35 & 0.874 & 0.138 \\
Learnable    & Direct xyz    & 24.62 & 0.802 & 0.204 \\
None  & Direct xyz  & 20.87 & 0.650 & 0.364 \\
\midrule
Patch depth  & Anchor $+$ offset & \textbf{28.46} & \textbf{0.894} & \textbf{0.118} \\
\bottomrule
\end{tabular}
\end{table}

%% file: Main/Partials/figure_uncertainty_vis.tex
\begin{figure}[t]
\centering
\includegraphics[width=1\linewidth]{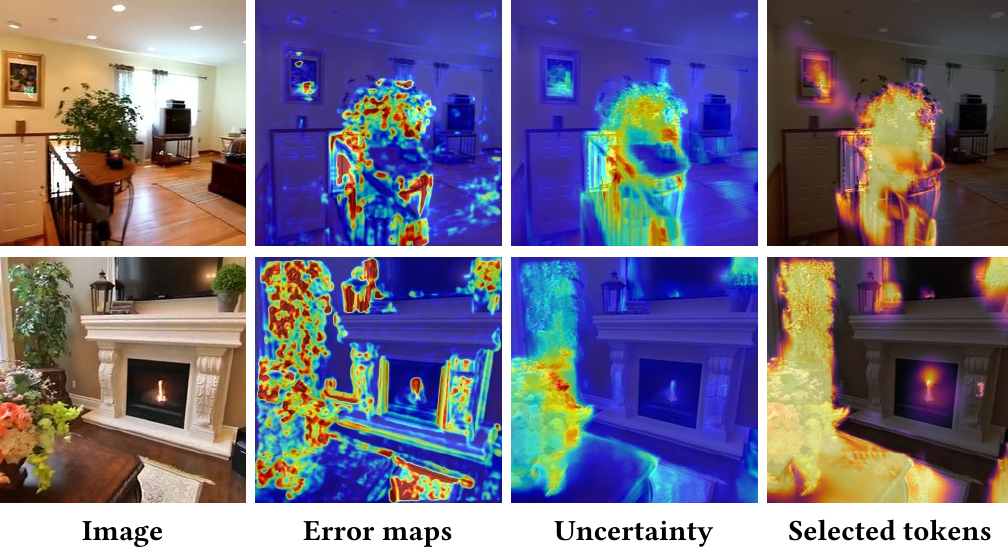}
\caption{
\textbf{Visualization of uncertainty scores and selected tokens} at the last block. We visualize actual error maps (D-SSIM), predicted uncertainty maps, and alpha maps of Gaussians decoded from the selected tokens.
}
\label{fig:uncertainty_vis}
\end{figure}

%% file: Main/Partials/table_ablation_ate.tex
\begin{table}[t]
\setlength{\tabcolsep}{4pt}
\centering
\small
\caption{\textbf{Ablation results on the ATE module.}}
\label{tab:ablation_ate}
\begin{tabular}{lcccc}
\toprule
Method & PSNR$\uparrow$ & SSIM$\uparrow$ & LPIPS$\downarrow$ \\
\midrule
Random & 27.97 & 0.888 & 0.126 \\
Farthest Point Sampling (FPS) & 27.98 & 0.888 & 0.125 \\
Fully learnable (STE)~\cite{ste}  & 27.93 & 0.887 & 0.125 \\
No expansion & 27.02 & 0.868 & 0.153 \\
Uncertainty & \textbf{28.46} & \textbf{0.894} & \textbf{0.118} \\
\bottomrule
\end{tabular}
\end{table}


%% file: Main/Sections/conclusion.tex
\section{Conclusion and Future Works}
We presented ATSplat, a feed-forward 3D Gaussian Splatting framework that restores the scene-adaptive capacity allocation of the 3DGS pipeline. 
Instead of densely assigning Gaussians to input image grids, ATSplat builds a sparse set of adaptive 3D anchor tokens, decodes them into a set of local Gaussians with 3D relative offsets, and progressively selects and expands tokens associated with challenging regions. 
This sparse-to-adaptive formulation decouples Gaussian placement from input pixels and allows us to allocate representational capacity according to scene complexity.
ATSplat reinterprets the core design principles of 3DGS within a feed-forward framework, achieving state-of-the-art quality with more than $5.7\times$ fewer Gaussians.

While ATSplat addresses the scene-adaptive capacity allocation problem in feed-forward 3DGS, several directions remain open for future work.
First, ATSplat expands tokens adaptively but does not prune those that become redundant during decoding.
A pruning mechanism analogous to that of 3DGS could be a key for improving decoder efficiency and more effective capacity allocation.
Second, several architectural improvements can be made for more scalable designs, to extend ATSplat to larger-scale scenes, more input views, and higher resolutions.
Finally, extending ATSplat to unposed settings would broaden its applicability to in-the-wild images.

%% file: Main/Sections/figure_only.tex
\input{Main/Partials/figure_qualitative_re10k}
\input{Main/Partials/figure_qualitative_dl3dv}

\input{Main/Partials/figure_qualitative_dl3dv_960p}

%% file: Main/Partials/figure_qualitative_re10k.tex
\begin{figure*}[t]
\centering
\includegraphics[width=1\linewidth]{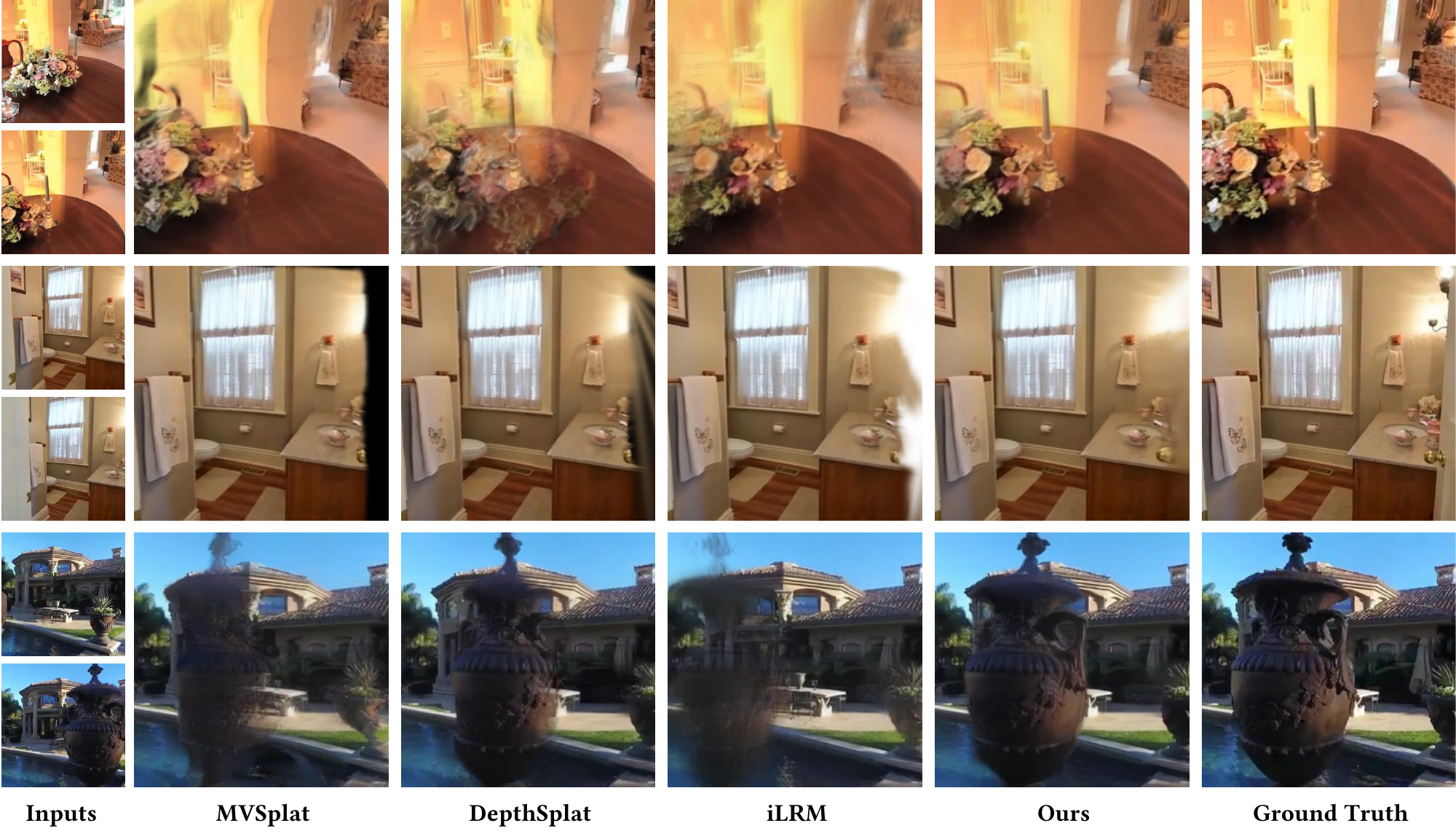}
\caption{
\textbf{Qualitative comparisons on RealEstate10K at $256\times256$ resolution} with 2 input views.
ATSplat reconstructs fine details in challenging regions while accurately reconstructing geometry, while using $5.7\times$ fewer Gaussians compared to the dense pixel-aligned baselines.
}
\label{fig:qualitative_re10k}
\end{figure*}

%% file: Main/Partials/figure_qualitative_dl3dv.tex
\begin{figure*}[t]
\centering
\includegraphics[width=1\linewidth]{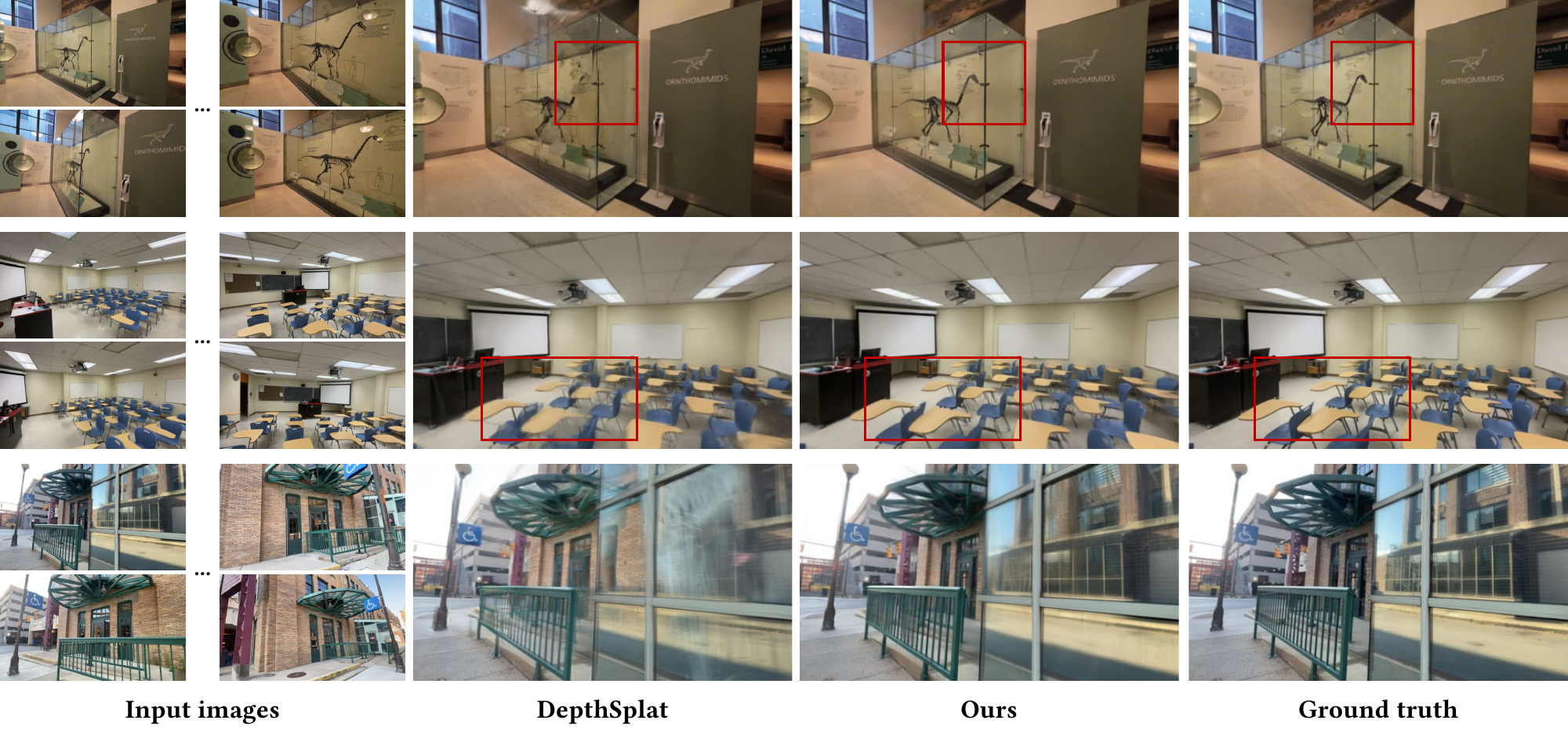}
\caption{
\textbf{Qualitative comparisons on DL3DV at $256\times448$ resolution} with 6 input views.
ATSplat produces high-fidelity renderings with only 120K Gaussians, which is 5.7$\times$ fewer than DepthSplat.
It also faithfully reconstructs thin structures, intricate details, and view-dependent appearance.
}
\label{fig:qualitative_dl3dv}
\end{figure*}

%% file: Main/Partials/figure_qualitative_dl3dv_960p.tex
\begin{figure*}[t]
\centering
\includegraphics[width=1\linewidth]{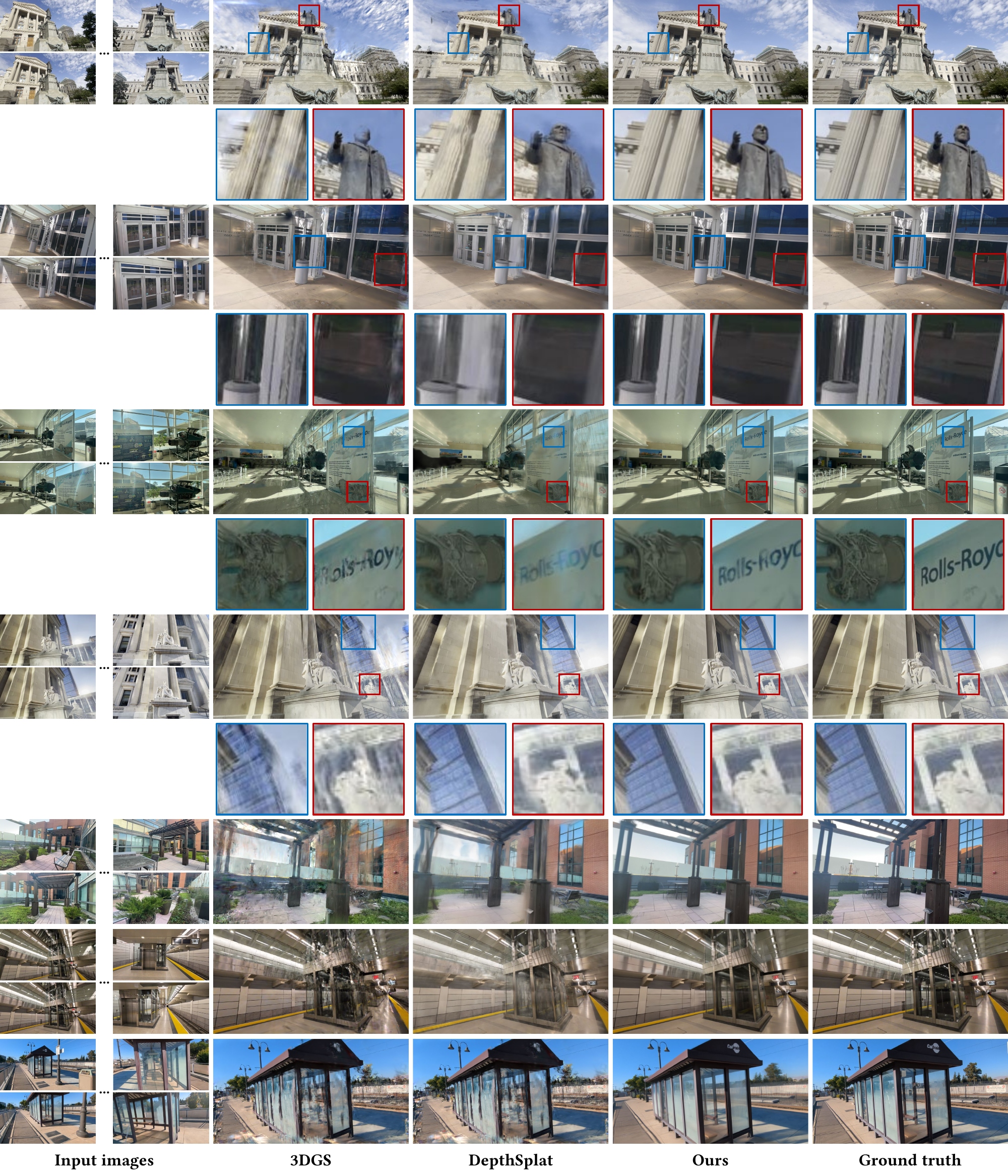}
\caption{
\textbf{Qualitative comparisons on DL3DV at $512\times 960$ resolution} with 12 input views.
ATSplat recovers fine details and thin structures in less than a second, with a significantly reduced number of Gaussians, whereas 3DGS requires over 10 minutes of per-scene optimization to obtain these results.
}
\label{fig:qualitative_dl3dv_960p}
\end{figure*}

%% file: Supple/supple.tex
\section{Additional Implementation Details}
In this section, we provide additional details on the model architecture and training configuration used in our experiments. These implementation details are included to facilitate reproducibility.

\subsection{Training Details}
We train our models with the PyTorch framework.
To improve memory efficiency during training, we employ the native scaled dot-product attention implementation provided by PyTorch. 
ATSplat is trained in BF16 precision, which reduces memory consumption while maintaining stable optimization in our experiments.
We optimize the model using AdamW~\cite{loshchilov2017adamw} with a weight decay of $0.05$. The initial learning rate is set to $2 \times 10^{-4}$ and is scheduled using cosine annealing after an initial warm-up phase.
We use a batch size of $8$ per GPU except for $256\times 448$ DL3DV~\cite{ling2024dl3dv} experiments, which results in a total batch size of $32$ for RealEstate10K~\cite{zhou2018re10k} and $64$ for DL3DV experiments.

\subsection{Architecture Details}
We use a feature dimension of $768$ throughout the model. 
The fine branch uses a patch size of $8$, while the coarse branch uses a patch size of $14$, following the resolution of DINO backbone. 
To ensure that the coarse and fine branches operate at a consistent relative resolution, we first downsample the input images before passing them to the DINO encoder in the coarse branch. This makes the spatial resolution of the fine patch grid twice that of the coarse patch grid.
Each decoder block leverages standard cross-attention layers.
Each of these layers consists of a cross-attention, a self-attention, and a feed-forward module.
We employ one such layer in the first two blocks, and two layers in the others.

In addition to DINO features, we additionally patchify the raw images using a patch size of $16$ and fuse the resulting image patch features to the corresponding DINO features through concatenation and a linear projection layer, after projecting them to the model dimension.
For the fine branch, before feeding to the decoder, we also inject the encoder output features as coarse-level context, into the fine-resolution patch features.
The final coarse patch features are upsampled to the spatial resolution of the fine patch grid and added to the fine patch features.

The coarse depth head and the Gaussian head are both implemented using two-layer MLPs.
Before each MLP head, we apply LayerNorm to stabilize the feature distribution. For the uncertainty head, we apply a softplus activation to the output.
We empirically found this parameterization to be more stable in our setup.
We also normalize the input camera poses before feeding them to the network.
Given the input camera sequence, we compute the average camera pose and transform the camera coordinate system such that the average pose is aligned with the identity pose, following previous works \cite{kang2025ilrm}.

\input{Supple/Partials/table_budget}
\subsection{Total Gaussian Budgets}
All results reported in the main paper are obtained with the fixed selection ratios $\rho_l$.
Under this setting, the number of tokens after the decoder is mainly determined by the input configuration.
Given $N_{\mathrm{init}}$ initial anchor tokens, the number of final tokens is
\begin{equation}
    N_{\mathrm{final}}
    \;=\;
    N_{\mathrm{init}} \prod_{l=1}^{L_{\mathrm{ATE}}} \bigl(1 + \rho_l (M - 1)\bigr),
\end{equation}
where $L_{\mathrm{ATE}}$ denotes the number of decoder blocks equipped with an ATE module and $M$ is the expansion ratio.
Each token is then decoded into $K$ local Gaussians, which yields $G = K N_{\mathrm{final}}$ primitives in total.
\tref{supp:tab:budget} lists the resulting token and Gaussian counts for every configuration used in our experiments.
We emphasize that fixed ratios are not the only choice at inference, and we further report per-scene dynamic budgets obtained by uncertainty thresholding in \sref{supp:sec:dynamic}.

\section{Additional Results}
\input{Supple/Partials/table_re10k_overlap}
\subsection{Results with Varying Viewpoint Overlaps}
\label{supp:sec:overlap}
We further evaluate ATSplat under varying input viewpoint overlaps, following the evaluation protocol of~\cite{ye2025noposplat}.
Small overlap typically corresponds to wider camera baselines, whereas large overlap corresponds to narrow baselines where most target viewpoints are placed close to the input views. 
\tref{supp:tab:re10k_overlap} reports PSNR, SSIM, and LPIPS on RealEstate10K across the resulting overlap bins.

ATSplat attains the strongest performance in the small-overlap regime by a clear margin, becomes comparable to pixel-aligned baselines under medium overlap, and eventually falls behind in the large-overlap setting.
This trend is a direct consequence of the differing formulations of ATSplat and pixel-aligned methods. 
Pixel-aligned methods predict per-pixel Gaussians anchored on input camera rays, and are therefore well-suited to large-overlap settings where most of the target content is directly observable from the input views.
As overlap decreases, however, an increasing portion of the target view lies away from the input rays, and ray-constrained placement becomes a fundamental limitation.
Our 3D anchor-offset effectively addresses this issue and achieves compelling results in the more challenging setup, thanks to our anchor-offset design that can place primitives freely in 3D space beyond input camera rays.

\input{Supple/Partials/table_yonosplat}
\subsection{Results with Square-Cropped Inputs}
We further compare ATSplat with YoNoSplat~\cite{ye2025yonosplat}, a concurrent feed-forward method evaluated on square-cropped images.
To match its configuration, we train and evaluate ATSplat on DL3DV at $224\times 224$ resolution with 6 input views, where both input and target images are square-cropped.
As shown in \tref{supp:tab:yonosplat}, ATSplat achieves higher quality across all metrics while using $4.1\times$ fewer Gaussians on average.

\input{Supple/Partials/table_cross_dataset}
\subsection{Cross-Dataset Generalization}
We next examine how our formulation transfers to unseen data.
We evaluate all models trained on RealEstate10K directly on DL3DV at $256\times 256$ resolution with 2 input views, without any fine-tuning.
\tref{supp:tab:cross_dataset} reports the zero-shot transfer results.
ATSplat attains the best PSNR and comparable SSIM and LPIPS while using significantly fewer Gaussians, indicating that ATSplat well adapts to unseen data distributions.

\input{Supple/Partials/table_extrapolation}
\subsection{Extrapolation}
Extrapolated views, which lie outside the input view window, pose a harder setting than the standard protocol, where target views are placed between the input views.
On DL3DV with 6 input views, we render the 10 frames placed before and after the input view window, so that a large portion of the target content is not directly observable from the input rays.
\tref{supp:tab:extrapolation} compares ATSplat with DepthSplat~\cite{xu2025depthsplat}.
This setting shares the same difficulty as the small-overlap regime in \sref{supp:sec:overlap}, where ray-constrained placement cannot cover content away from the input rays.
ATSplat adaptively places primitives freely in 3D space, and thus achieves better results as the target viewpoints move away from the input views.

\input{Supple/Partials/table_unified}
\subsection{A Unified Model across Datasets and View Counts}
Following previous feed-forward methods~\cite{charatan2024pixelsplat, chen2024mvsplat, xu2025depthsplat}, we train a separate model for each resolution.
This choice follows the common evaluation protocol for fair comparisons and is not a property of ATSplat, as none of its components is tied to a specific resolution or number of input views.
To verify this, we train a single model jointly on both datasets across all input view counts, and evaluate it under each setup.
\tref{supp:tab:unified} compares the unified model with the separately trained ones.
The unified model achieves comparable quality with the separately trained models, verifying that ATSplat can be trained jointly across datasets and view counts.

\section{Additional Analysis and Ablations}

\input{Supple/Partials/table_dynamic_budget}
\input{Supple/Partials/figure_decoder_expansion}
\subsection{Dynamic Gaussian Budgets via Uncertainty Thresholds}
\label{supp:sec:dynamic}
With fixed selection ratios, ATSplat distributes primitives adaptively within a budget that is determined by the input configuration, rather than selecting the budget itself.
The same model can instead expand every token whose uncertainty exceeds certain thresholds, which produces scene-dependent budgets at inference without retraining or architectural changes.

As shown in \tref{supp:tab:dynamic_budget}, the resulting budget varies by up to $6.9\times$ across scenes, indicating that it now follows scene complexity rather than the input configuration.
Lower thresholds expand more tokens and improve quality, and only the smallest reaches the state-of-the-art quality, still using $3.4\times$ fewer Gaussians compared to iLRM.

\input{Supple/Partials/table_k_scaling}
\subsection{Scaling the Gaussian Budget}
The number of Gaussians decoded from each anchor token, $K$, can control the total Gaussian budget of ATSplat.
We vary $K$ from $4$ to $64$ on RealEstate10K with 2 input views, keeping all other settings identical.
As shown in \tref{supp:tab:k_scaling}, rendering quality improves consistently as $K$ grows, with diminishing returns beyond our default setting of $K=16$.

\subsection{Expansion at each decoder block}
To gain deeper insights into the ATE module, we further analyze its behavior at each decoder block by progressively enabling expansion from the earliest block to the latest, while keeping all other settings identical.
\fref{supp:fig:decoder_expansion} shows the resulting renderings.
With expansion fully disabled, the limited representation capacity leaves some fine details missing.
As expansion is enabled at additional blocks, the renderings are progressively refined, with fine-grained texture and high-frequency detail recovered more faithfully.

\input{Supple/Partials/table_ablation_anchor_dl3dv}
\input{Supple/Partials/table_ablation_ate_dl3dv}
\subsection{Ablations on DL3DV}
The ablations in the main paper are conducted on RealEstate10K with 2 input views.
We further validate our core components on the DL3DV dataset with 6 input views, to examine whether they remain effective on larger and more complex scenes.
\tref{supp:tab:ablation_anchor_dl3dv} and \tref{supp:tab:ablation_ate_dl3dv} report the results for the 3D anchor designs and the ATE module, respectively.
On the DL3DV dataset, regressing absolute Gaussian centers without any anchor initialization fails to converge, indicating that coarse 3D anchors become more critical as the spatial extent of the scene grows.

\section{Additional Qualitative Results}
We provide additional qualitative comparisons on RealEstate10K and DL3DV. 
\fref{supp:fig:more_qualitative_re10k} presents results on RealEstate10K.
\fref{supp:fig:more_qualitative_dl3dv} and \fref{supp:fig:more_qualitative_dl3dv_960} present results on the DL3DV dataset at $256\times448$ and $512\times960$ resolutions, respectively. 
We further refer readers to the supplementary video for more comprehensive results, which include rendered sequences of ATSplat and side-by-side comparisons.

\clearpage
\input{Supple/Partials/figure_more_qualitative_re10k}
\input{Supple/Partials/figure_more_qualitative_dl3dv}
\input{Supple/Partials/figure_more_qualitative_dl3dv_960}

%% file: Supple/Partials/table_budget.tex
\begin{table}[t]
\setlength{\tabcolsep}{3pt}
\centering
\footnotesize
\caption{
\textbf{Token and Gaussian counts} under our experimental configurations.
Each anchor token is decoded into $K=16$ Gaussians, and the total number of Gaussians is $G = K N_{\mathrm{final}}$.
*: the first ATE block uses a selection ratio of $0.8$ instead of $0.5$ in this setting.
}
\label{supp:tab:budget}
\begin{tabular}{llcccc}
\toprule
Dataset & Resolution & Views & $N_{\mathrm{init}}$ & $N_{\mathrm{final}}$ & $G$ \\
\midrule
RE10K & $256\times 256$ & 2  & 512  & 1440  & 23040 \\
DL3DV & $256\times 448$ & 2  & 896  & 2520  & 40320 \\
DL3DV & $256\times 448$ & 4  & 1792 & 5040  & 80640 \\
DL3DV & $256\times 448$ & 6  & 2688 & 7560  & 120960 \\
DL3DV & $512\times 960$* & 12 & 5760 & 19440 & 311040 \\
\bottomrule
\end{tabular}
\end{table}

%% file: Supple/Partials/table_re10k_overlap.tex
\begin{table*}
\small
\setlength{\tabcolsep}{3pt}
\centering
\caption{\textbf{Quantitative comparisons on RE10K} with varying viewpoint overlaps. Smaller viewpoint overlaps indicate larger camera baseline ranges.}
\label{supp:tab:re10k_overlap}
\begin{tabular}{l c ccc ccc ccc ccc}
\toprule
& & \multicolumn{3}{c}{Small} & \multicolumn{3}{c}{Medium} & \multicolumn{3}{c}{Large} & \multicolumn{3}{c}{Average} \\
\cmidrule(lr){3-5} \cmidrule(lr){6-8} \cmidrule(lr){9-11} \cmidrule(lr){12-14}
Method & \#Gauss.$\downarrow$
       & PSNR$\uparrow$ & SSIM$\uparrow$ & LPIPS$\downarrow$
       & PSNR$\uparrow$ & SSIM$\uparrow$ & LPIPS$\downarrow$
       & PSNR$\uparrow$ & SSIM$\uparrow$ & LPIPS$\downarrow$
       & PSNR$\uparrow$ & SSIM$\uparrow$ & LPIPS$\downarrow$ \\
\midrule
pixelSplat~\cite{charatan2024pixelsplat} 
& 131K & 20.28 & 0.719 & 0.265 & 23.73 & 0.811 & 0.180 & 27.15 & 0.880 & 0.121 & 23.86 & 0.808 & 0.184 \\
MVSplat~\cite{chen2024mvsplat}          
& 131K & 20.37 & 0.725 & 0.250 & 23.81 & 0.814 & 0.172 & 27.47 & 0.885 & 0.115 & 24.01 & 0.812 & 0.175 \\
DepthSplat~\cite{xu2025depthsplat}       
& 131K & 22.82 & 0.798 & 0.193 & 25.38 & 0.851 & 0.145 & 28.32 & \underline{0.900} & \underline{0.104} & 25.59 & 0.852 & \underline{0.145} \\
iLRM~\cite{kang2025ilrm} 
& 131K & \underline{23.82} & \underline{0.813} & \textbf{0.184} & \underline{26.54} & \textbf{0.864} & \textbf{0.139} & \textbf{29.43} & \textbf{0.910} & \textbf{0.103} & \textbf{26.70} & \textbf{0.864} & \textbf{0.140} \\
Ours
& \textbf{23K} & \textbf{24.09} & \textbf{0.815} & \underline{0.185} & \textbf{26.56} & \underline{0.861} & \underline{0.144} & 29.13 & \underline{0.900} & 0.114 & \textbf{26.70} & \underline{0.861} & 0.146 \\
\bottomrule
\end{tabular}
\end{table*}

%% file: Supple/Partials/table_yonosplat.tex
\begin{table}[t]
\setlength{\tabcolsep}{4pt}
\centering
\footnotesize
\caption{
\textbf{Quantitative comparisons on DL3DV at $224\times 224$ resolution} with 6 input views.
*: the number of Gaussians is averaged over the test scenes due to the opacity-based Gaussian pruning.
}
\label{supp:tab:yonosplat}
\begin{tabular}{lcccc}
\toprule
Method & PSNR$\uparrow$ & SSIM$\uparrow$ & LPIPS$\downarrow$ & \#Gauss.$\downarrow$ \\
\midrule
YoNoSplat~\cite{ye2025yonosplat} & 24.72 & 0.817 & 0.139 & 212K* \\
Ours & \textbf{27.97} & \textbf{0.882} & \textbf{0.118} & \textbf{52K} \\
\bottomrule
\end{tabular}
\end{table}

%% file: Supple/Partials/table_cross_dataset.tex
\begin{table}[t]
\setlength{\tabcolsep}{4pt}
\centering
\footnotesize
\caption{
\textbf{Zero-shot transfer from RealEstate10K to DL3DV} at $256\times 256$ resolution with 2 input views.
All methods are trained on RealEstate10K and evaluated on DL3DV without fine-tuning.
}
\label{supp:tab:cross_dataset}
\begin{tabular}{lcccc}
\toprule
Method & PSNR$\uparrow$ & SSIM$\uparrow$ & LPIPS$\downarrow$ & \#Gauss.$\downarrow$ \\
\midrule
MVSplat~\cite{chen2024mvsplat} & 16.18 & 0.459 & 0.441 & 131K \\
DepthSplat~\cite{xu2025depthsplat} & 18.07 & 0.571 & \textbf{0.352} & 131K \\
iLRM~\cite{kang2025ilrm} & \underline{18.82} & \textbf{0.588} & \textbf{0.352} & 131K \\
Ours & \textbf{19.17} & \underline{0.584} & \underline{0.371} & \textbf{23K} \\
\bottomrule
\end{tabular}
\end{table}

%% file: Supple/Partials/table_extrapolation.tex
\begin{table}[t]
\setlength{\tabcolsep}{6pt}
\centering
\small
\caption{
\textbf{Extrapolated novel-view synthesis on DL3DV} with 6 input views.
Target views are sampled from the 10 frames placed before and after the input view window.
}
\label{supp:tab:extrapolation}
\begin{tabular}{lccc}
\toprule
Method & PSNR$\uparrow$ & SSIM$\uparrow$ & LPIPS$\downarrow$ \\
\midrule
DepthSplat~\cite{xu2025depthsplat} & 19.22 & 0.687 & 0.265 \\
Ours & \textbf{22.11} & \textbf{0.747} & \textbf{0.246} \\
\bottomrule
\end{tabular}
\end{table}

%% file: Supple/Partials/table_unified.tex
\begin{table}[t]
\setlength{\tabcolsep}{3pt}
\centering
\footnotesize
\caption{
\textbf{Comparison of separately trained models and a single unified model} jointly trained across both datasets and all input view counts.
RE10K is evaluated at $256\times 256$ and DL3DV at $256\times 448$ resolution.
}
\label{supp:tab:unified}
\begin{tabular}{lccc ccc}
\toprule
& \multicolumn{3}{c}{Separate} & \multicolumn{3}{c}{Unified} \\
\cmidrule(lr){2-4} \cmidrule(lr){5-7}
Setting
 & PSNR$\uparrow$ & SSIM$\uparrow$ & LPIPS$\downarrow$
 & PSNR$\uparrow$ & SSIM$\uparrow$ & LPIPS$\downarrow$ \\
\midrule
RE10K, 2 views & 28.46 & 0.894 & 0.118 & 28.11 & 0.890 & 0.125 \\
DL3DV, 2 views & 20.07 & 0.630 & 0.332 & 21.80 & 0.702 & 0.283 \\
DL3DV, 4 views & 25.67 & 0.827 & 0.172 & 25.73 & 0.830 & 0.169 \\
DL3DV, 6 views & 27.28 & 0.868 & 0.138 & 27.26 & 0.868 & 0.136 \\
\bottomrule
\end{tabular}
\end{table}

%% file: Supple/Partials/table_dynamic_budget.tex
\begin{table}[t]
\setlength{\tabcolsep}{3pt}
\centering
\footnotesize
\caption{
\textbf{Per-scene dynamic budgets on RealEstate10K} with 2 input views.
Each threshold triplet lists the uncertainty thresholds applied at the three ATE blocks, replacing the fixed selection ratios at inference.
}
\label{supp:tab:dynamic_budget}
\begin{tabular}{lccc ccc}
\toprule
& & & & \multicolumn{3}{c}{\#Gauss.$\downarrow$} \\
\cmidrule(lr){5-7}
Selection & PSNR$\uparrow$ & SSIM$\uparrow$ & LPIPS$\downarrow$ & Avg. & Min. & Max. \\
\midrule
Fixed ratios      & 28.46 & 0.894 & 0.118 & 23K & 23K & 23K \\
\midrule
0.1 / 0.1 / 0.4   & 27.90 & 0.888 & 0.127 & 13K & 8K  & 36K \\
0.05 / 0.05 / 0.2 & 28.35 & 0.895 & 0.118 & 19K & 8K  & 55K \\
0.01 / 0.01 / 0.04 & 28.64 & 0.899 & 0.112 & 39K & 14K & 65K \\
\bottomrule
\end{tabular}
\end{table}

%% file: Supple/Partials/figure_decoder_expansion.tex
\begin{figure*}[t]
\centering
\includegraphics[width=1\linewidth]{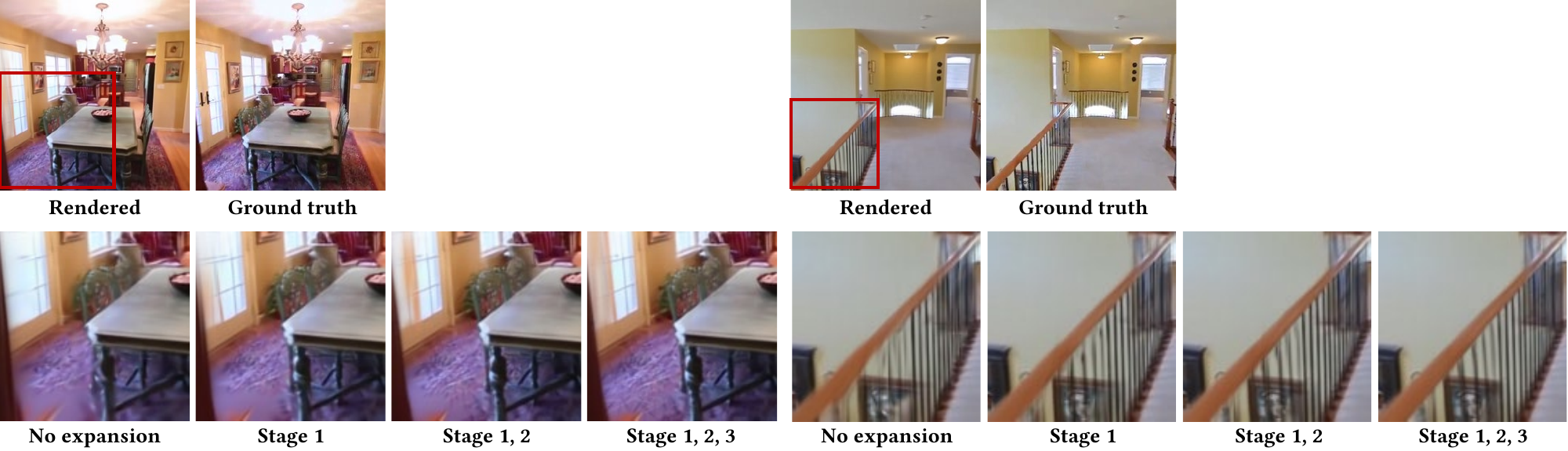}
\caption{
\textbf{Effect of progressively enabling token expansion across decoder blocks.}
Starting from a configuration with expansion disabled at all decoder blocks, we gradually enable expansion from the earliest block to the latest. As more blocks are activated, fine-grained texture and thin structures are progressively recovered.
}
\label{supp:fig:decoder_expansion}
\end{figure*}

%% file: Supple/Partials/table_k_scaling.tex
\begin{table}[t]
\setlength{\tabcolsep}{6pt}
\centering
\small
\caption{
\textbf{Effect of the number of Gaussians per anchor token $K$} on RealEstate10K with 2 input views.
*: our default setting used in all other experiments.
}
\label{supp:tab:k_scaling}
\begin{tabular}{lcccc}
\toprule
$K$ & PSNR$\uparrow$ & SSIM$\uparrow$ & LPIPS$\downarrow$ & \#Gauss.$\downarrow$ \\
\midrule
4    & 27.63 & 0.877 & 0.141 & 5K \\
8    & 28.25 & 0.890 & 0.124 & 11K \\
16*  & 28.46 & 0.894 & 0.118 & 23K \\
32   & 28.66 & 0.898 & 0.113 & 46K \\
64   & 28.73 & 0.899 & 0.111 & 92K \\
\bottomrule
\end{tabular}
\end{table}

%% file: Supple/Partials/table_ablation_anchor_dl3dv.tex
\begin{table}[t]
\setlength{\tabcolsep}{7pt}
\centering
\small
\caption{\textbf{Ablation results on the 3D anchor designs} on DL3DV with 6 input views.}
\label{supp:tab:ablation_anchor_dl3dv}
\begin{tabular}{llccc}
\toprule
Anchor init. & Gaussian center & PSNR$\uparrow$ & SSIM$\uparrow$ & LPIPS$\downarrow$ \\
\midrule
None         & Ray $+$ depth & 26.37 & 0.845 & 0.161 \\
Learnable    & Direct xyz    & 21.05 & 0.625 & 0.359 \\
None         & Direct xyz    & \multicolumn{3}{c}{Diverged} \\
\midrule
Patch depth  & Anchor $+$ offset & \textbf{27.28} & \textbf{0.868} & \textbf{0.138} \\
\bottomrule
\end{tabular}
\end{table}

%% file: Supple/Partials/table_ablation_ate_dl3dv.tex
\begin{table}[t]
\setlength{\tabcolsep}{4pt}
\centering
\small
\caption{\textbf{Ablation results on the ATE module} on DL3DV with 6 input views.}
\label{supp:tab:ablation_ate_dl3dv}
\begin{tabular}{lccc}
\toprule
Method & PSNR$\uparrow$ & SSIM$\uparrow$ & LPIPS$\downarrow$ \\
\midrule
Random & 26.31 & 0.846 & 0.154 \\
Farthest point sampling (FPS) & 26.21 & 0.843 & 0.158 \\
Fully learnable (STE)~\cite{ste} & 26.72 & 0.859 & 0.143 \\
No expansion & 25.17 & 0.810 & 0.200 \\
Uncertainty & \textbf{27.28} & \textbf{0.868} & \textbf{0.138} \\
\bottomrule
\end{tabular}
\end{table}

%% file: Supple/Partials/figure_more_qualitative_re10k.tex
\begin{figure*}[t]
\centering
\includegraphics[width=1\linewidth]{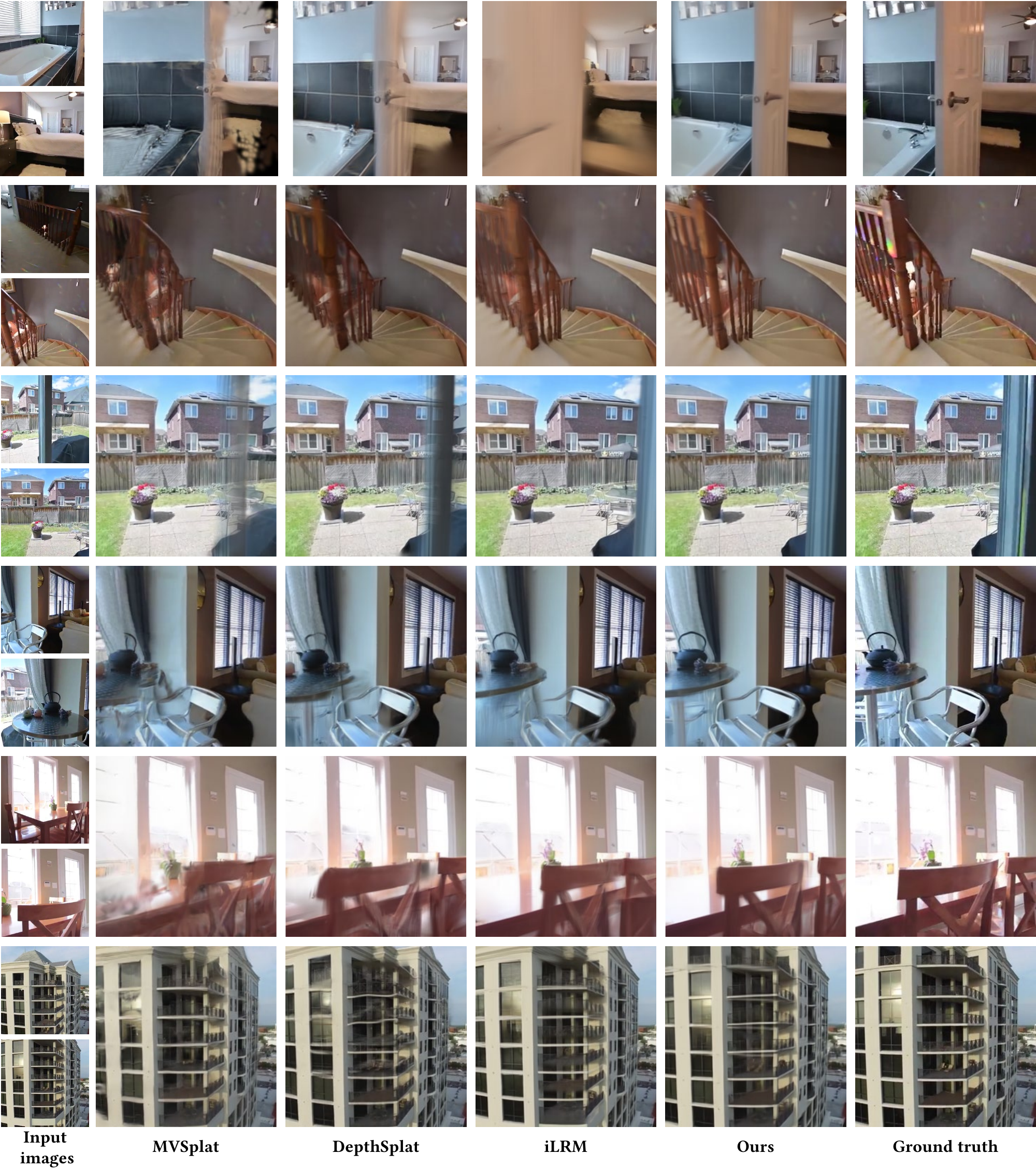}
\caption{
\textbf{Additional qualitative comparisons on RealEstate10K at $256\times256$ resolution} with 2 input views.
}
\label{supp:fig:more_qualitative_re10k}
\end{figure*}

%% file: Supple/Partials/figure_more_qualitative_dl3dv.tex
\begin{figure*}[t]
\centering
\includegraphics[width=1\linewidth]{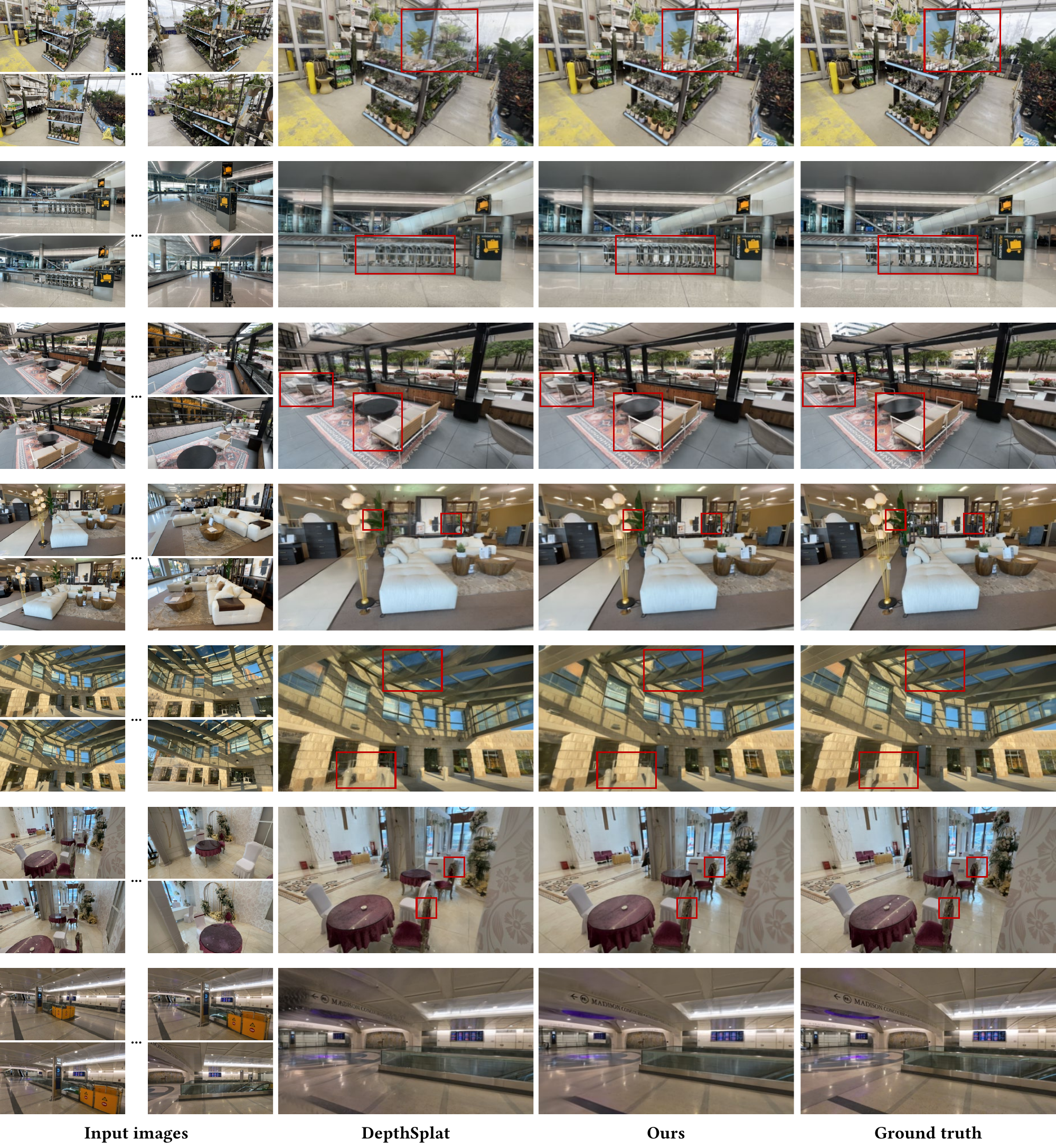}
\caption{
\textbf{Additional Qualitative comparisons on DL3DV at $256\times448$ resolution} with 6 input views.
}
\label{supp:fig:more_qualitative_dl3dv}
\end{figure*}

%% file: Supple/Partials/figure_more_qualitative_dl3dv_960.tex
\begin{figure*}[t]
\centering
\includegraphics[width=1\linewidth]{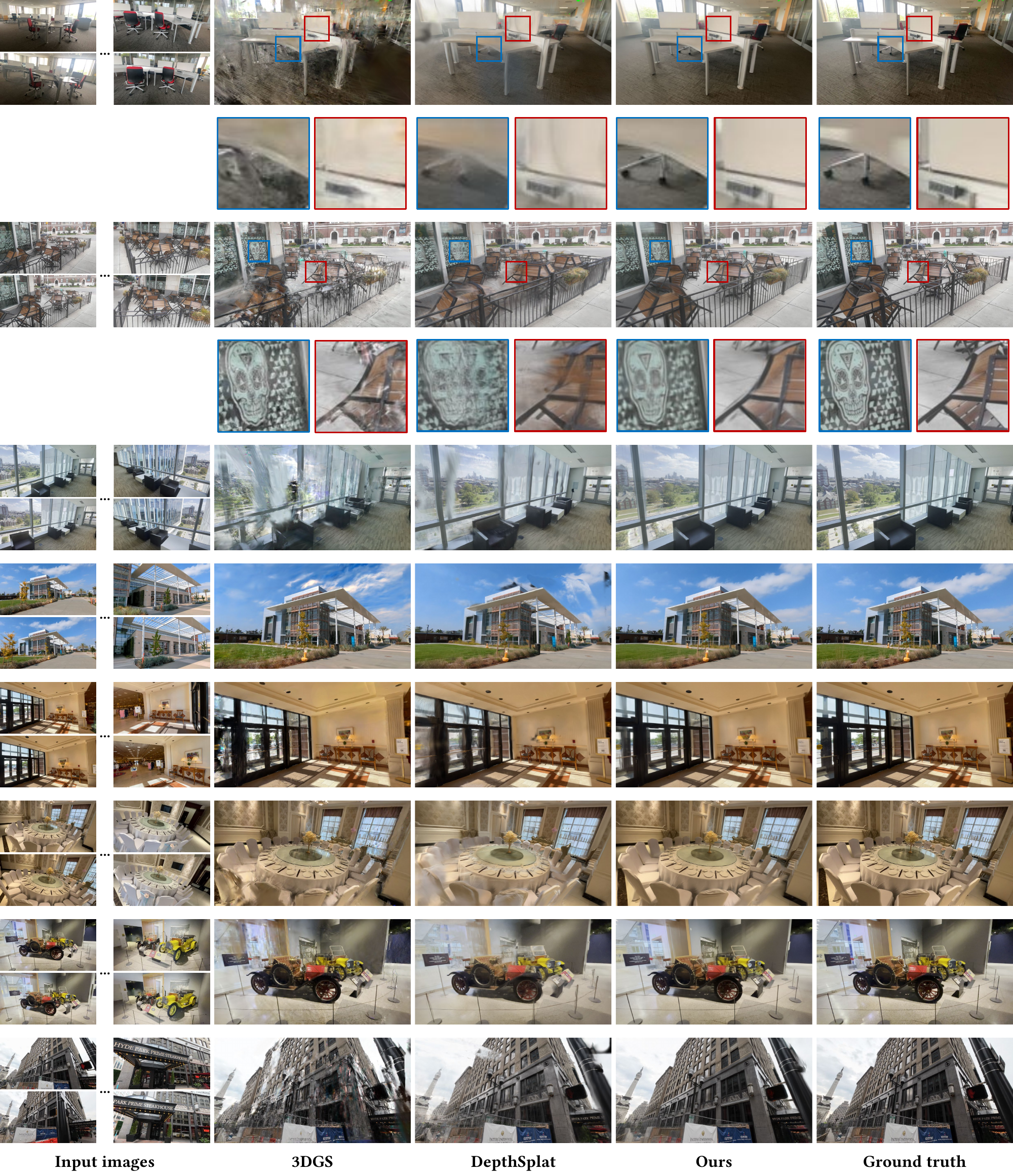}
\caption{
\textbf{Additional Qualitative comparisons on DL3DV at $512\times960$ resolution} with 12 input views.
}
\label{supp:fig:more_qualitative_dl3dv_960}
\end{figure*}